\documentclass{clv3}

\usepackage{xcolor}
\usepackage{multirow}
\usepackage{booktabs}
\usepackage{amsmath}
\usepackage{graphicx}
\usepackage{tabularx}
\usepackage{subcaption}
\usepackage{algorithm}
\usepackage{algorithmic}
\usepackage{url}

\begin{document}
\issue{1}{1}{2016}


\runningtitle{Taxonomical Semantic Parsing}
\runningauthor{Xiao Zhang, Gosse Bouma and Johan Bos}

\title{Neural Semantic Parsing with Extremely Rich Symbolic Meaning Representations}

\author{Xiao Zhang}
\affil{University of Groningen \\
Center for Language and Cognition\\
\textit{xiao.zhang@rug.nl}}

\author{Gosse Bouma}
\affil{University of Groningen \\
Center for Language and Cognition\\
\textit{g.bouma@rug.nl}}

\author{Johan Bos}
\affil{University of Groningen \\
Center for Language and Cognition\\
\textit{johan.bos@rug.nl}}

\maketitle

\begin{abstract}
Current open-domain neural semantics parsers show impressive
performance. However, closer inspection of the symbolic meaning representations they produce reveals significant weaknesses: sometimes they tend to merely copy character sequences from the source text to form symbolic concepts, defaulting to the most frequent word sense based in the training distribution.
By leveraging the hierarchical structure of a lexical ontology, we introduce a novel compositional symbolic representation for concepts based on their position in the taxonomical hierarchy. This representation provides richer semantic information and enhances interpretability.
We introduce a neural "taxonomical" semantic parser  to utilize this new representation system of predicates, and compare it with a standard neural semantic parser trained on the traditional meaning representation format, employing
a novel challenge set and evaluation metric for evaluation. 
Our experimental findings demonstrate that the taxonomical model, trained on much richer and complex meaning representations, is slightly subordinate in performance to the traditional model using the standard metrics for evaluation, but outperforms it when dealing with out-of-vocabulary concepts. We further showed through neural model probing that training on a taxonomic representation enhances the model's ability to learn the taxonomical hierarchy.
This finding is encouraging for research in computational semantics that aims to combine data-driven distributional meanings with knowledge-based symbolic representations.
\end{abstract}

\section{Introduction} \label{sec:intro}

The task of formal semantic parsing is to map natural language expressions (words, sentences and texts) to unambiguous interpretable formal meaning representations. 
The components of these meaning representations can be divided into two parts: the domain-independent logical symbols (such as negation $\lnot$, conjunction $\land$, disjunction $\lor$, equality $=$, and the quantifiers $\exists$, $\forall$), and the non-logical symbols (the concepts and relations between them---the predicates), possibly tailored to a specific domain. It is the latter component that forms the focus of this article, as we see several shortcomings in the way predicates are represented in mainstream computational semantics, since this is usually done by combining a lemma, part-of-speech tag and sense number.
This representation of concepts is not only language-specific, but also doesn't exploit the power of pre-trained language models used in neural approaches, currently the state of the art in semantic parsing \cite{bai-etal-2022-graph,martinez-lorenzo-etal-2022,wang-etal-2023-pre}.

\begin{figure}[hbtp]
    \centering
    \includegraphics[width=\columnwidth]{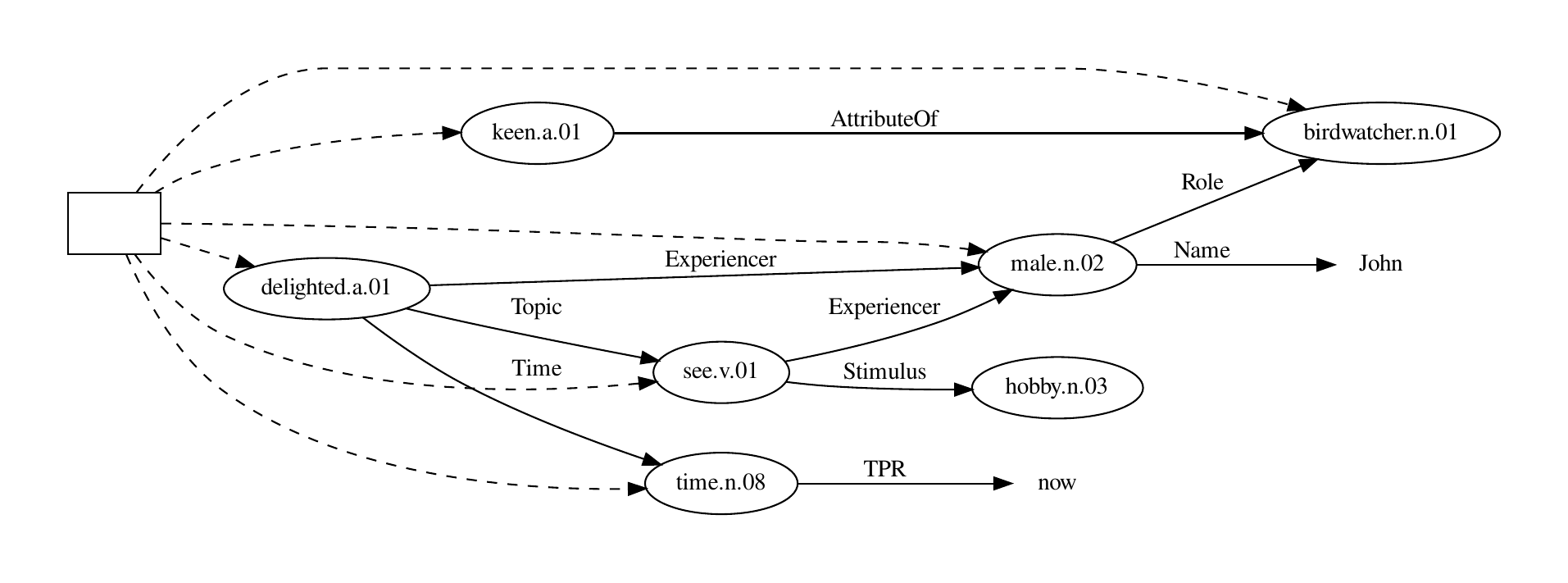}
    \caption{Graphical display of a meaning representation for the sentence \emph{John, a keen birdwatcher, was delighted to see a hobby} in the style of the Parallel Meaning Bank. Oval nodes represent concepts, boxed nodes introduce contexts, and labeled edges denote thematic roles or semantic relations.}
    \label{fig:standard}
\end{figure}

We illustrate this issue with an example. Figure~\ref{fig:standard} shows a meaning representation in the form of a directed acyclic graph, as is common in currently used frameworks in computational semantics \cite{allen-etal-2008-deep,amr,oepen-etal-2020-mrp,Abzianidze2020CONNL,Bos2023IWCS}. The nodes represent concepts that are symbolized by the lemmas of the content words (nouns, verbs, adjectives, adverbs) that triggered them. As content words are often polysemous, they are usually disambiguated using one of the standard lexical ontologies, including WordNet \citep{Fellbaum-1998-wordnet}, OntoNotes \citep{hovy-etal-2006-ontonotes}, VerbNet \citep{Bonial-2011-verbnet} and BabelNet \citep{Navigli-2012-BabelNet}. This can be done by adding a part-of-speech sign and sense number suffix to the symbol, and is mainstream practice across a wide spectrum of semantic formalisms, such as AMR \citep{amr}, BMR \citep{martinez-lorenzo-etal-2022}, DRS \citep{gmb}, PTG \cite{hajic-etal-2012-announcing}, EDS \cite{oepen-lonning-2006-discriminant}. However, the (interrelated) problems that we observe with this approach are the following:

\begin{itemize}

\item The representations of concepts are usually not normalized: they are represented in various ways based on synonym lemmas of the same language or translated lemmas for other languages;

\item The predicate symbols possess minimal or no inherent semantic structure. This makes it impossible to determine how they are related to other predicates without access to external knowledge bases;

\item The predicate symbols are essentially atomic. Yet, a neural network's tokenizer will typically break down the symbols into meaningless sequences of characters to reduce the size of its vocabulary.

\end{itemize}

The first problem can be illustrated with a simple example. Consider the English synonyms \emph{car} and \emph{automobile}. In most current approaches, their concepts would be represented with different predicate symbols (e.g., \emph{car.n.01} and \emph{automobile.n.01}, following WordNet), even though they share the same lexical meaning. For non-English lexicalizations for this concept, for instance Italian \emph{macchina} or French \emph{voiture}, one could opt to use English-based predicate symbols (as is done by \citet{pmb} or use a multi-lingual lexical ontology \cite{martinez-lorenzo-etal-2022}, but these solutions don't apply any kind of normalization: the same meaning is represented by different predicate symbols.

The second problem has risen more to the foreground with the rise of distributional semantics to capture word meanings. While certain predicate symbols exhibit an accidental internal structure that provides additional insight about them (e.g., \emph{blackbird.n.01} is a kind of \emph{bird.n.01}), for most symbols, it is not possible to determine, without external resources, that \emph{dog.n.01} is closely related to and compatible with \emph{animal.n.01} or \emph{puppy.n.01}. This is in stark contrast with semantics based on embeddings where meanings are represented by large vectors based on their contextual occurrences in corpora \cite{mikolov}. Some attempts have already been made to close the gap between semantic networks and semantic spaces \cite{rothe-schutze-2015-autoextend,saedi-etal-2018-wordnet,Scarlini2020SensEmBERTCS,lees-etal-2020-embedding}.

The third problem is perhaps more of a theoretical issue, with the desire to use a clean and sound methodology for composing meanings. From an engineering point of view, what happens  under the hood of a semantic parser isn't that important as long as it produces satisfactory accuracy scores.  
However, a clear disadvantage is that predicted concepts are not guaranteed to exist in the external knowledge base (e.g., WordNet). 

\smallskip

These three problems were not particularly pressing in the past or considered problematic, perhaps mainly because word sense disambiguation was seen as a separate task in the sequential symbolic pipeline of traditional semantic parsing. However, the advent of neural methods in semantic processing has magnified these concerns. Neural networks, due to their statistical nature, are confined to their training data distribution, and are therefore struggling in understanding and generating out-of-distribution concepts \cite{johnson2017clevr, Lake2017GeneralizationWS, kim-linzen-2020-cogs, li-etal-2023-slog, groschwitz-etal-2023-amr, zhang-etal-2024-gaining}. Additionally, neural networks tend to find typographical correspondences between the input words and output concepts \cite{edman2024characterlevel}, thereby copying from the input which, we argue, artificially inflates parsing performance scores. Nevertheless, we believe that pre-trained language models possess the capability to comprehend the semantics of unseen words and concepts within context, but the current representations used in semantic parsing form an obstacle to exploit it. 

The following scenario illustrates this issue.
Consider a neural semantic parser trained on a corpus that pairs English sentences with meanings that encode concepts in the standard symbolic format based on a lemma and a sense number.
We give this parser the sentence in Figure~\ref{fig:standard} about birdwatcher John who spotted a hobby, a rare type of falcon. Assume that the word \emph{hobby} is not in the training data at all. It is very likely that this kind of parser would associate the word \emph{hobby} with the incorrect concept \emph{hobby.n.01}, the "spare-time activity" sense, because it learned how to lemmatize from other examples and the suffix "n.01" appears to be the most frequent noun sense. By chance, the parser could produce the correct concept \emph{hobby.n.03} (the bird sense), maybe because the suffix "n.03" is often associated with lemmas that end in "by". In our view, we don't want to develop semantic parsers that make a \emph{wild guess} for unknown concepts. 
Instead, we would like to develop semantic parsers that make an \emph{educated guess}: perhaps it could come up with a concept that it has seen during training, e.g., \emph{bird.n.01} or even \emph{falcon.n.01} that appears in a similar context as the sentence it tries to parse. In this article, we introduce and evaluate a method for integrating pre-trained language models with symbolic meaning representations that utilize taxonomical encodings.

\section{Taxonomical Encodings}\label{sec:tax-encoding-intro}

\begin{figure}[hbtp]
    \centering
    \includegraphics[height=70mm]{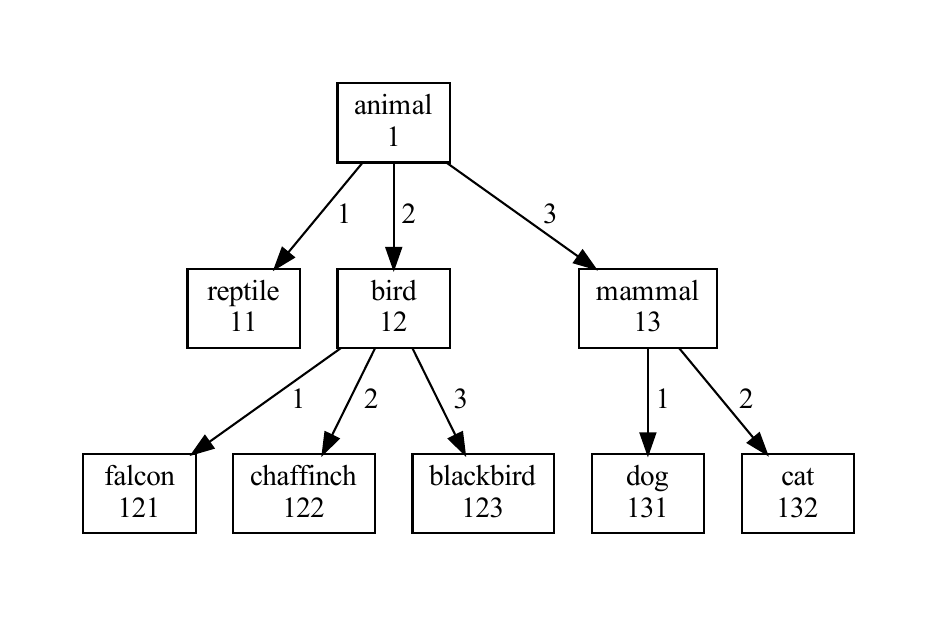}
    \caption{Simple illustration of a taxonomical encoding.  Each box denotes a concept within a typical ontological ISA-hierarchy. The longer the encoding, the more specific its concepts is.}
    \label{fig:theidea}
\end{figure}

In the approach we follow, we radically deviate from the view of representing non-logical symbols based on lemmas, part-of-speech tags and sense numbers. Instead we introduce a way to encode concepts and relations with internally structured blocks of meaning based on external lexical ontologies or taxonomies. These predicate symbols are not directly interpretable for human beings as they do not correspond to units of language. For instance, the taxonomical encodings for \emph{cat} and \emph{dog} share a common prefix that tells us that both are of the category \emph{mammal}. The encodings for adjectives like \emph{good} and \emph{bad} will be very similar, with the only difference indicating that they are antonyms. The idea, illustrated in Figure~\ref{fig:theidea}, draws inspiration from methods in image classification \cite{mukherjee,makingbettermistakes}.

Our research questions are the following. 
First of all, how can we take existing lexical ontologies as a starting point to encode symbols for concepts (nouns, verbs, adjectives, adverbs) in a compositional way? 
Secondly, are current neural models actually capable of learning these abstract symbols that have no direct connection with any surface form of language, without  reducing parsing performance too much? Furthermore, how can we intuitively assess the model's understanding of the hierarchy of concepts, and does the model trained on taxonomical encodings demonstrate a superior performance?
Finally, do taxonomical encodings of concepts give the semantic parser the ability to make sense of out-of-distribution concepts by assigning a meaning that is compatible with or close to the ground truth, rather than producing a symbol based on the lemma of the word and hoping for the best. In other words, can the parser make an \emph{educated guess} based on distributional meaning rather than on superficial features when it encounters word meanings that it has never seen before, taking advantage of pre-trained language models?

Our work's primary contribution lies in presenting a taxonomy-based approach to meaning representation, which is very distinct from prior methods. Accordingly, we connect symbolic and neural techniques in a semantic parsing system. It is \emph{symbolic} because the semantic parser generates interpretable and transparent meaning representations.
It is \emph{neural} because it harnesses the power of distributional meaning models and neural networks to enhance robustness in unexpected circumstances. 
A direct consequence of our new approach to semantic parsing is that current evaluation methods, that are based on graph matching, are not sufficient for our purposes, as comparing concepts will be beyond simple string matching and needs to be based and implemented using semantic rather than syntactic similarity. 
Hence, to assess the capabilities to deal with out-of-vocabulary concepts, we have developed a new challenge test and new semantic evaluation tools. Furthermore, we probe the embeddings of concepts within neural models to evaluate the encoded taxonomical information.
Our experiments reveal that our new representation-based parser performs comparably to the original representation in standard tests but excels in predicting unknown concepts on challenge sets. Additionally, the probing tests indicate that neural models trained with the new representation learn more taxonomical information compared to the original representation.

\section{Background and Related Work\label{sec:background}}

\subsection{Semantic Parsing}

Early approaches to semantic parsing were mostly based on rule-based systems with compositional semantics defined on top of a syntactic structure obtained by a parser \cite{woods1973progress,Hendrix1977DevelopingAN,templeton-burger-1983-problems}. 
The development of syntactic treebanks contributed to the creation of robust statistical syntactic parsers, thereby facilitating semantic parsing with broad coverage \cite{bos-etal-2004-wide}. The emergence of neural methodologies and the availability of extensive semantically annotated datasets \cite{amr, gmb, pmb} marked a shift in semantic parsing techniques, diminishing the emphasis on syntactic analysis \cite{barzdins-gosko-2016-riga, noord-2017-neural, Bevilacqua2021OneST}. The introduction of pre-trained language models within the sequence-to-sequence framework led to further enhancements in parsing accuracy \cite{samuel-straka-2020-ufal, shou-lin-2021-incorporating, Lee2021MaximumBS, Bevilacqua2021OneST, lee-etal-2022-maximum, bai-etal-2022-graph, martinez-lorenzo-etal-2022, wang-etal-2023-pre}.

Various meaning representations have been the target for semantic parsing---for excellent recent overviews, see \citet{conll-2020-conll} and \citet{sadeddine-etal-2024-survey}.
Dominating the field of semantic parsing is Abstract Meaning Representation, AMR, facilitated by the large supply of annotated data and the simplicity of its meaning structures, that are based on simple directed acyclic graphs. There are various extensions proposed to enrich AMR. These are, among others, BMR (Babelnet Meaning Representation), that incorporates multi-lingual semantic resources \cite{martinez-lorenzo-etal-2022}, 
and UMR (Uniform Meaning Representation), that includes discourse level phenomena \cite{umr}.
In this work we focus on DRS, a richer meaning representation, a semantic formalism that drew substantial interest in computational semantics \cite{bos-etal-2004-wide, evang-bos-2016-cross, rik-tacl-nural-2018, van-noord-etal-2019-linguistic, evang-2019-transition, liu-etal-2019-discourse-representation, van-noord-etal-2020-character, wang-etal-2021-input, poelman-etal-2022-transparent, wang-etal-2023-pre, zhang-etal-2024-gaining}.

\subsection{Evaluating Semantic Parsing}\label{subsec:evaluation}

A common way of evaluating semantic parsing is to compare parser output with a gold standard using graph overlap \cite{allen-etal-2008-deep,smatch,counter}.
A well-known implementation is Smatch, \emph{Semantic Match},
that measures the structural similarity by mapping graphs into triples and determining the maximum amount of triples that are shared by two semantic graphs, computed by taking the harmonic mean of precision and recall of matching triples \cite{smatch}. Smatch is mostly used to assess AMR parsing, but not exclusively, and \citet{poelman-etal-2022-transparent} adopt it for DRS parsing.
The original Smatch implements matching of semantic material without further nuances, so several variations and extensions of Smatch have been  developed to enhance semantic evaluation \cite{damonte-etal-2017-incremental, cai-lam-2019-core, Opitz-2020-s2match, wein-schneider-2022-accounting, opitz-2023-smatch}.

\citet{cai-lam-2019-core} argue that more weight should be given to triples that form the core of the meaning expressed by a semantic graph.
They propose a variant of Smatch that takes root distance into account by reducing the significance of triple matches that are further away from the root of the graph \cite{cai-lam-2019-core}. In the meaning representations of our choice, DRS, there is no designated root, so we cannot adopt this variant of Smatch straightaway. Although we think the idea of giving more importance to triples that form the core of the meaning expressed by a text is interesting, we believe more research is required to establish what exactly constitutes this and we therefore will consider this outside the scope of this article. 

\citet{Opitz-2020-s2match} argue that the "hard" matching of Smatch is not always justified and propose
S$^2$Match, \emph{Soft Similarity Match}, by implementing a graded semantic match of concepts 
with the help of a distance function that computes a number between $0$ and $1$. 
The distance function can be anything that fulfills its purpose. 
\citet{Opitz-2020-s2match}  showcase their idea using GloVe embeddings to get a similarity score between lemmas.   
\citet{wein-schneider-2022-accounting} propose LaBSE embeddings \cite{feng-etal-2022-language} to compute the similarity between two concepts in different languages for cross-lingual comparison of meaning representations.
However, these choices of distance functions ignore scenarios where different concepts are expressed by the same lemma. We embrace the use of S$^2$Match in our work but will replace the distance function by an operation that measures the ontological distance between two concepts or relations.

\subsection{WordNet} \label{subsec:wordnet}

We make heavy use of the lexical ontology WordNet, a handcrafted electronic dictionary \cite{Fellbaum-1998-wordnet}.
In WordNet, words are organized around \emph{synsets}, i.e., sets of words that have similar meanings. A synset consists of one or more words; an ambiguous word (a word with more than one \emph{sense}) is placed into several synsets, one for each distinct meaning. Synsets are connected to each other by several semantic relations (see below).
Each synset has a unique identifier, a 9-digit number based on the byte offset in the WordNet database, where the first number identifies the part of speech.

The original WordNet  was designed for English \cite{Fellbaum-1998-wordnet}. Subsequent efforts have been undertaken to establish WordNets for various languages and to develop multilingual lexical resources \cite{Navigli-2012-BabelNet, Vossen1998IntroductionTE,bond2013linking}, or to include WordNet into a formal ontology \cite{gangemi} and to integrate it with Wikipedia \cite{YAGO,Speer201ConceptNet5.5}. 

Only parts of speech of content words are included in WordNet: nouns (n), verbs (v), adjectives (a), and adverbs (r).
There are several ways to refer to a specific synset: by using the unique identifier, or more commonly, by a combination of lemma, part of speech, and sense number of one of its members.
For instance, the English noun \emph{hobby} is placed into three synsets in Princeton's WordNet for English: the first sense, \emph{hobby.n.01}, is glossed as "an auxiliary activity", and two of its other synset members are \emph{pursuit.n.03} and \emph{avocation.n.01}; \emph{hobby.n.02} denotes the sense of plaything for children, and \emph{hobby.n.03} refers to the bird of prey with the scientific name \emph{Falco subbuteo}.

The power of WordNet manifests itself by the relations between synsets that it offers. The hypernym and hyponym relations connect generic with specific synsets.
For instance, the direct hypernym of \emph{hobby.n.03} is \emph{falcon.n.01}. Synsets that share the same hypernym are called co-hyponyms. For example, \emph{hobby.n.03} and \emph{peregrine.n.01} are co-hyponyms, as they are both falcons. Verbs are similarly organised in WordNet, but the term \emph{troponym} is used instead of hyponym to indicate a more fine-grained sense of a verb. For instance, \emph{falcon.v.01} is a troponym of \emph{hunt.v.01}. Some verbs are related to other verbs via the \emph{entailment} relation, e.g., \emph{oversleep.v.01} entails \emph{sleep.v.01}, and some verbs
have derivationally related forms corresponding to nouns, e.g., \emph{hunt.v.01} is related to \emph{hunt.n.08}.

Adjectives and adverbs are quite differently inserted into WordNet than nouns and verbs are. Adjectives are arranged by falling into one of the categories of \emph{head} and \emph{satellite}, the former playing a more pivotal role, and the latter a specialization of a certain adjective. For some adjectives, there exists the \emph{antonym} relation between synsets of head adjectives, indicating opposite meanings. For instance, \emph{good.a.01} and \emph{bad.a.01} are antonyms, with satellites \emph{cracking.a.01}, \emph{superb.a.02}, among others, for the former, and \emph{awful.a.01} and \emph{terrible.a.02} (among many others) for the latter. Some adjectives are connected to attribute nouns, e.g., fast.a.01 has the attribute synset speed.n.02. 
Adjectives are also connected to their derivationally related forms of verbs and nouns. Some adverbs are connected to adjective by the \emph{pertainym} relation, e.g., \emph{quickly.r.01} is a pertainym of \emph{quick.a.01}. However, there are several adverbs that have no connection with other synsets in WordNet.

The structure of WordNet triggered various proposals to calculate some kind of similarity score between two concepts. \citet{Resnik1995UsingIC} proposed a similarity metric based on the notion of information content, which requires an external corpus to calculate the the frequencies of concepts. The Leacock-Chodorow similarity calculates similarity based on the hypernym/hyponym path length between synsets \cite{Leacock1998}. \citet{Wu-1994-wup} propose a similar way to compute the conceptual distance between synsets, but include both the depth of the concepts in the WordNet hierarchy and their least common subsumer (LCS, e.g., the first hypernym they share). The Wu-Palmer Score (WPS, Equation~\ref{equ:wps}), is  the metric we adopt because it is easy to implement and  independent of the depth of the hierarchy.

\begin{equation}
    \textrm{WPS} = 2 * \frac{depth(\textrm{LCS}(s_1, s_2))}{depth(s_1) + depth(s_2)}
    \label{equ:wps}
\end{equation}

For instance, the WPS of the first and third (semantically unrelated) senses of the noun \emph{hobby} is low: WPS(hobby.n.01, hobby.n.03)=0.087, 
whereas the similarity between hobby (the bird) and falcon is high:
WPS(falcon.n.01, hobby.n.03)=0.963.
Note that the similarity metrics mentioned are grounded in WordNet's taxonomy and typically are most appropriate for noun comparisons, and less so for other parts of speech. In Section~\ref{sec:tax}, we delve into the distinct taxonomies of verbs, adjectives, and adverbs in WordNet. 
These structural differences lead to inaccuracies in measuring concept similarity for non-noun categories using the existing WordNet taxonomy. To address this issue, we calculate the similarity using the taxonomy encodings described in Section~\ref{sec:tax} instead of using the original WordNet taxonomy.

One could also view our taxonomical encoding as a vector, and apply cosine similarity for assessing the similarity between two concepts. However, we won't get the nuances that we need if we follow this approach. This is because cosine similarity would
not take the position of the element within the vector into account, thereby overlooking the inherent hierarchical structure of our taxonomical encodings. For instance, considering Figure~\ref{fig:theidea}, the cosine similarity for "falcon" and "chaffinch" would be the same as that for "falcon" and "dog", and this is not what we would expect, as the former pair of concepts is more similar than the latter.

\subsection{Non-logical Symbols, Concepts, and Word Senses}

Symbolic meaning representations consist of the logical and non-logical parts. The non-logical symbols, the predicates and relations, define the concepts in the domain of interest. 
Here we assume an open-domain approach where nouns, verbs, adjectives and adverbs are mapped to predicates taken from an ontology,
proper names\footnote{Named entity linking and grounding is an important part of semantic processing, but is outside the scope of this article and not relevant for meeting our research objectives.} and numbers are mapped to literals,
and prepositions and implicit arguments are mapped to an inventory of roles and relations. 

In NLP,  distinct formats for non-logical symbols (also known as \emph{predicate symbols}) have been adopted, ranging from words, lemmas, a combination of a lemma and a sense number, to entries in a lexical ontology. 
In AMR \cite{kingsbury-palmer-2002-treebank}, predicate symbols are only partially disambiguated. 
Some symbols are derived from PropBank framesets and formatted as lemma-sense (e.g., see-01) but most predicates do not include senses and are simply represented by their corresponding lemma.
BMR \cite{martinez-lorenzo-etal-2022} follows the graph structure of AMR but the predicates are encoded by leveraging the multilingual semantic network of BabelNet, interpreting the non-logical symbols from WordNet, Wikipedia, and other resources.
In the DRS representation of the Parallel Meaning Bank \cite[PMB]{pmb}, predicates for noun, verbs, adjectives and adverbs follow the corresponding WordNet synsets and adhere strictly to the lemma-pos-sense format (see Figure~\ref{fig:standard}).

Hence, one important subtask of semantic parsing is Word Sense Disambiguation (WSD), the process of identifying the appropriate meaning of a word within its context. Traditionally, this is approached as a classification task with the goal of selecting the correct sense from a set of predefined sense inventories \cite{ijcai2021p593}. In contrast, the task of concept prediction in semantic parsing is treating WSD as a generative task. Here, semantic parsers are expected to generate the correct word sense without being given the inventories of senses. This presents a significant challenge, as it is considered currently “impossible” to accurately generate word senses without external knowledge sources, particularly when the word senses have not been encountered in the training data \cite{groschwitz-etal-2023-amr}.

In the case of semantic parsing, the sense number system acts as an instrument for sense prediction. However, numbering senses is fairly arbitrary, only constrained by the tendency that senses encountered frequently in corpora get assigned a low sense number. 
Consequently, for concepts not encountered in the training data, predicting whether a word corresponds to sense number 3, 4 or 5 holds no distinguishable difference: the best strategy to use for a WSD component would be choosing a low sense number, like 1 or 2. Our new representation for concepts presented in Section~\ref{sec:tax} addresses this issue by refraining from using sense numbers and instead incorporating taxonomical information in concept representation.

\subsection{Sense Embeddings}
\label{sec:sense_emb}

The aim of this  article is to develop and evaluate a new way of representing concepts and incorporating them into formal meaning representations of natural language sentences. Another way of representing concepts are \emph{sense embeddings}, pre-trained vectors extracted from a neural model, usually based on language models trained on a corpus with labeled word senses.
Various approaches have been introduced to generate sense embeddings. 
AutoExtend \cite{rothe-schutze-2015-autoextend} derives synset and lexeme embeddings from word embeddings. 
Context-AwaRe Embeddings of Senses \cite{scarlini-etal-2020-ares} uses a semi-supervised approach to producing sense embeddings for the lexical meanings within a lexical knowledge base.
SensEmBERT \cite{Scarlini2020SensEmBERTCS} employs an approach by combing the power of the language modeling and the knowledge contained in a semantic network. 
Pre-trained sense embeddings are known to improve word sense disambiguation \cite{oele-noord-2018-simple,bevilacqua-navigli-2020-breaking}. 

However, due to their size and floating point numbers, pre-trained sense embeddings are (obviously) not directly appropriate to be explicitly part of a formal meaning representation such as the one shown in Figure~\ref{fig:standard}. One option would be to compress the embeddings \cite{Andrews2015CompressingWE} and transform the numbers into integers, but this solution falls outside the objectives of this paper.
Hence, although we will not explore the use of pre-trained sense embeddings to improve semantic parsing, in Section~\ref{sec:experiments} we will compare them
with the sense embeddings extracted from the semantic parsing models that we develop to get an idea how well they reflect the taxonomical hierarchy encoded by WordNet.

\subsection{Discourse Representation Structures} \label{sec:sbn}

In order to run our experiments we need a reasonably-sized annotated corpus of sentences and their meaning representations. 
Several such annotated corpora are available for AMR, Abstract Meaning Representation \cite{amr}, and the majority of current semantic
parsing methods are developed using AMR datasets.
However, for our purposes, we are not able to make use of these linguistic resources because only
part of the non-logical symbols (predicates) are disambiguated in AMR, as we outlined in the previous section.\footnote{A quick calculation on the AMR 2017 corpus revealed that about 60\% of the predicates are not sense-disambiguated. Most of these are predicates for nouns.}

Instead, we will work with a variant of Discourse Representation Structure (DRS), the meaning representation proposed in Discourse Representation Theory \cite{kamp-1993-discourse}. The Parallel Meaning Bank (PMB) offers a large corpus of sentences paired with DRSs with concepts represented by WordNet synsets and a neo-Davidsonian event semantics with VerbNet-inspired thematic roles. 

\begin{figure}[hbtp]
\centering 
\small
\begin{tabular}{|l|}
 \hline
     x$_1$ x$_2$ x$_3$ s$_1$ s$_2$ e$_1$ x$_4$\\
     \hline
     male.n.02(x$_1$) \ Name(x$_1$, "John")\\
     keen.a.01(s$_1$) \ AttributeOf(s$_1$,x$_3$)\\
     person.n.01(x$_2$) \ x$_2$=x$_1$ \ Role(x$_2$,x$_3$) \\ birdwatcher.n.01(x$_3$)\\
     delighted.a.01(s$_2$) \ Experiencer(s$_2$,x$_1$) \ Topic(s$_2$,e$_1$)\\
  see.v.01(e$_1$) \ Experiencer(e$_1$,x$_1$) \ Stimulus(e$_1$,x$_4$) \\
   hobby.n.03(x$_4$)\\
     \hline
\end{tabular}
\hfill
\begin{tabular}{ll}
male.n.02 Name "John"\\
keen.a.01 AttributeOf +2\\
person.n.01 EQU -2 Role +1 \\
birdwatcher.n.01\\
delighted.a.01 Experiencer -3 Topic +1\\
see.v.01 Experiencer -1 Stimulus +1\\
hobby.n.03\\
\end{tabular}
\caption{Discourse Representation Structure for a sentence shown in box format (left) and sequence notation (right). The corresponding graph for this DRS is shown in Figure~\ref{fig:standard}.}\label{fig:drs}
\end{figure}

The formal language of DRS consists of discourse referents,
and DRS conditions. DRS conditions are predicates applied to discourse referents, relations between discourse referents or literals, or comparison statements (i.e., equality, approximately, temporal precedence, see Appendix). DRSs are recursive data structures; complex DRSs can be constructed to express negation, conjunction, and discourse relations.

An example DRS in box format, the equivalent for the meaning representation graph in Figure~\ref{fig:standard}, is shown on the left in Figure~\ref{fig:drs}. However, in our experiments in Section~\ref{sec:experiments}, we will use neither of these formats when training our semantic parsing models. Instead, we will use the sequence notation for DRS where variables are replaced by De Bruijnian indices \cite{Bos2023IWCS}. The sequence notation is a convenient way for training neural semantic parsers that are based on seq2seq architectures, because there are variables names and a minimal amount of punctuation symbols.
Our running example in sequence notation is shown on the right in Figure~\ref{fig:drs}.

\section{Encoding Concepts and Relations}\label{sec:tax}

There are four parts of speech in WordNet, all with a different ontological organization. Therefore, we describe for each category how we compute its taxonomical encodings. These encodings will be our new way to represent concepts in a formal meaning representation and used to improve semantic parsing. We use Princeton WordNet version 3.0 \cite{Fellbaum-1998-wordnet}, compatible with the Parallel Meaning Bank.

\subsection{Nouns} 

For the encoding of nouns we will make use of the WordNet hyponym-hypernym relation between synsets. Each noun synset has one or more hypernyms, except entity.n.01, which therefore represents the most general synset.
For noun synsets with more than one synset (i.e., indicating multiple inheritance) we consider just one of the possible hypernyms. 

This procedure  maps all the noun synsets to one large isa-hierarchy with the top node \emph{entity.n.01}. Given a synset within this obtained hierarchy, we give each direct hyponym-hypernym edge a label (a single ASCII character, excluding the zero "0"), ensuring that the labels for each co-hyponyms are all distinct\footnote{In cases where concepts exhibit multiple inheritances, we choose the first hypernym path.}. Once we have done this for each edge, we can read off a unique sequence of labels for each synset. The maximum length of this sequence is the maximum depth of hyponym-hypernym links in WordNet. We pad encodings with trailing zeroes in order to give each synset encoding the same length\footnote{Initial experiments comparing with padded and non-padded encodings reveal that models with padding outperform those without. We also experimented with pruning the encodings, removing nodes in the hierarchy that are non-branching and do not appear in the data. But this idea also yielded worse results.}
The number of different labels that we need corresponds to the maximum of co-hyponyms for a synset in WordNet. 

Table~\ref{tab:hyponyms} gives a snapshot of how this labeling works. The resulting taxonomical code gives us a symbolic representation that groups similar concepts (i.e., synsets) together based on their internal structure. The more labels that they share from left-to-right in the encoding, the more they have in common. The more zeroes an encoding has, the more general its synsets is. To distinguish noun synsets from other parts of speech, we attach the prefix "n" to the encoding.

\begin{table}[hbtp]
\caption{Snapshot of generated taxonomy encodings for a group of related noun synsets. Synonyms receive the same encodings, hyponyms get encodings that are more specific (less zeroes).}\label{tab:hyponyms}
\begin{tabular}{lcc}
\toprule
\textbf{WordNet synset member}       & \textbf{WordNet ID} & \textbf{Taxonomical encoding}       \\ 
\midrule
entity.n.01           & 100001740           & n$1000000000000000000000$ \\
food.n.01             & 100021265           & n$1233100000000000000000$ \\
beverage.n.01         & 107881800           & n$1233110000000000000000$ \\
drink.n.03            & 107881800           & n$1233110000000000000000$ \\
alcoholic\_drink.n.01 & 107884567           & n$1233111000000000000000$ \\
alcohol.n.01          & 107884567           & n$1233111000000000000000$ \\
brew.n.01             & 107886572           & n$1233111100000000000000$ \\
beer.n.01             & 107886849           & n$1233111110000000000000$ \\
booze.n.01            & 107901587           & n$1233111200000000000000$ \\
brandy.n.01           & 107903208           & n$1233111210000000000000$ \\
\bottomrule
\end{tabular}
\end{table}

\subsection{Verbs} 

For verb synsets we follow essentially the same procedure as for nouns presented in tne previous section making use of the troponym and entailment relations available in WordNet. However, the hierarchy of verbs is much flatter than that of nouns, resulting in too many top nodes (synsets without hypernym).  For verb synsets without a hypernym, we create edges to noun synset to which they are derivationally related, as shown in Table~\ref{tab:verb}. For noun synsets inserted in this way, we expand the hierarchy as we did for nouns. To distinguish verb concepts from noun-derived concepts, we attach the prefix "v" to it.

\begin{table}[hbtp]
\caption{Snapshot of generated taxonomy encodings that link verb synsets to a  noun synset.}
\label{tab:verb}
\begin{tabular}{lcc}
\toprule
\textbf{WordNet synset member} & \textbf{WordNet Id} & \textbf{Taxonomical encoding}       \\ 
\midrule
change\_of\_integrity.n.01 & 100376063 &n$1111211500000000000000$ \\
separation.n.09            & 100383606 &n$1111211510000000000000$ \\
removal.n.01               & 100391599 &n$1111211511000000000000$ \\
get\_rid\_of.v.01          & 202224055 &v$1111211511100000000000$ \\
throw\_away.v.01           & 202222318 &v$1111211511110000000000$ \\
abandon.v.01               & 202228031 &v$1111211511111000000000$ \\
dump.v.02                  & 202224945 &v$1111211511120000000000$ \\
\bottomrule
\end{tabular}
\end{table}

\subsection{Adjectives and Adverbs}\label{sec:adj&adv}

For adjective synsets we create a hierarchical link between satellites and their heads. The head adjectives will be related to noun synsets using derivationally related verb or noun synsets or attribute nouns. To distinguish the adjective encodings, we attach the prefix "a" to it. Antonyms receive the same encodings but are decorated with a positive or negative suffix\footnote{When calculating the Wu-Palmer similarity between adjectives/adverbs, this suffix is moved to the end of the last non-zero character.}. 
WordNet doesn't provide information whether an antonym is positive or negative, so we use a simple heuristic to check the prefix of the adjective's lemma (im-, non-, un-), see \citet{van-son-etal-2016-building, blanco-moldovan-2010-automatic}. Adverbs are linked to adjectives via the pertainym relation and receive the same encoding but with the prefix "r".

\begin{table}[hbtp]
\caption{Snapshot of generated taxonomy encodings that link adjective and adverb synsets to a noun synset. Encodings for adjectives and adverbs have an additional suffix encoding  polarity.}
\label{tab:adj}
\begin{tabular}{lcl}
\toprule
\textbf{WordNet synset member} & \textbf{WordNet Id} & \textbf{Taxonomical encoding}       \\ 
\midrule
speed.n.02  & 105058140 & \texttt{n}$1133A31000000000000000$ \\
fast.a.01   & 300976508 & \texttt{a}$1133A31100000000000000$$+$ \\
fast.r.01   & 400086000 & \texttt{r}$1133A31100000000000000$$+$ \\
lazy.a.01   & 300981304 & \texttt{a}$1133A31121000000000000$$-$ \\
slow.a.01   & 300980527 & \texttt{a}$1133A31120000000000000$$-$  \\
slowly.r.01 & 400161630 & \texttt{r}$1133A31120000000000000$$-$ \\
quick.a.01  & 300979366 & \texttt{a}$1133A31130000000000000$$+$ \\
haste.n.01      & 105060189 & \texttt{n}$1133A31300000000000000$ \\
abruptness.n.03 & 105060476 & \texttt{n}$1133A31310000000000000$ \\
sudden.a.01   & 301143279 & \texttt{a}$1133A31311000000000000$ $|$ \\
all\_of\_a\_sudden.r.02 & 400061677 & \texttt{r}$1133A31311000000000000$ $|$ \\
suddenly.r.01 & 400061677 & \texttt{r}$1133A31311000000000000$ $|$ \\
\bottomrule
\end{tabular}
\end{table}

\subsection{Roles, Operators and Discourse Relations}

The meaning representations that we employ follow a neo-Davidsonian way of representing events \cite{Parsons1990-PAREIT}, where events are related to their participants by a close set of thematic roles, i.e., Agent, Theme, Patient, Result, and so on. 
The inventory of roles is an extension of the hierarchical set proposed in VerbNet \cite{Bonial-2011-verbnet}, extended with roles used in the Parallel Meaning Bank. The elaboration of the complete taxonomy of these roles is meticulously outlined in Appendix~\ref{app:roles}. There are also roles to connect non-event entities, for instance those appearing in genitive constructions or noun compounds. Some thematic roles are paired with their inverse roles, for instance, \emph{Sub} and \emph{SubOf}. To clearly distinguish between these roles, we employ distinct prefixes: 't' denotes a thematic role, whereas 'i' signifies an inverse role.
We also convert each Discourse Relation and operator into distinct mathematical symbols, as shown in Appendix~\ref{app:operators}. In contrast to roles, these two logical components lack a taxonomy structure, therefore their encoding is straightforward, involving a direct mapping to single-byte symbols.

Now that we have explained how the encoding process works for concepts, roles, operators and discourse relations, we can  put everything together and transform the semantic graph into a graph encoded with the WordNet Identifier and WordNet taxonomical encoding, as illustrated in the graphs in Appendix~\ref{app:graphformat}. 
The sequential representations of meanings, which are used for training in Section~\ref{sec:experiments}, are illustrated in Table~\ref{tab:seq}.

\begin{table}[!htbp]
\caption{First-order Logic representation and the three sequential meaning representations (Lemma-Pos-Sense, WordNet-Identifier and Taxonomical-encodings) 
for "John doesn't laugh."}
\centering
\begin{tabular}{ll}
\toprule
FOL & $\exists$x(male.n.02(x) $\land$ Name(x,"John") $\land$ $\exists$t(time.n.08(t) $\land$ t=now\\
     & $\lnot\exists$e(laugh.v.01(e) $\land$ Agent(e,x) $\land$ Time(e,t)))\\
     \hline
LPS & male.n.02 Name "John"       time.n.08 EQU now \\
    & NEGATION <1 laugh.v.01 Agent --2 Time --1\\
    \hline
WID & 109624168 500000018 "John" 115135822 = now\\
     &    $\neg$ <1    200031820 500000004 --2 500000003 --1\\
     \hline
TAX & n1222211P00000000000000 t12000 "John" n1133222000000000000000 = now\\
    &    $\neg$ <1     v2B20000000000000000000 t22100 --2 t21000 --1\\
    \bottomrule
\end{tabular}
\label{tab:seq}
\end{table}

\section{Developing Taxonomy-based Semantic Tools}\label{sec:eval}

Several new tools are required to work with the taxonomical encoding of concepts that we proposed in the previous section.
First of all,  we  need to revise the existing way of measuring semantic parsing performance. We will do so by replacing the well-known Smatch metric with one that takes concept similarity into account. Second, we need a new challenge set that measures parsing performance on out-of-distribution concepts. Finally, we need to add an interpretation component to the semantic parser that maps taxonomical encodings back to human-readable concepts.

\subsection{Soft Semantic Matching using the WordNet Taxonomy}

We adopt the S$^2$Match framework of \citet{Opitz-2020-s2match} (see Section~\ref{subsec:evaluation}) 
but replace its distance function to incorporate taxonomical encodings.  Recall that Smatch converts a semantic graph into node-edge-node triples and computes a score based on the maximum number of matching triples.
In standard Smatch, two triples get a matching score of 1 if and only if there is a perfect match between the two nodes and the edge. 
S$^2$Match extends this approach by introducing a soft matching between instance triples, 
where a distance function based GloVe embedding similarity returns a score between 0 and 1. We modify Smatch and S$^2$Match with respect to three issues:

\begin{enumerate}

\item We replace the distance function based on word embeddings by the Wu-Palmer Score (see Section~\ref{subsec:wordnet});

\item We not only allow soft matching based on the Wu-Palmer Score for instance triples, but also for role triples;

\item In the implementation of Smatch, triples featuring TOP were discarded since DRS, unlike AMR, does not contain roots.

\end{enumerate}

In the rest of this article we refer to these alterations of Smatch and S$^2$Match as Hard Smatch and Soft Smatch, respectively. Note that the  Wu-Palmer distance based on the standard WordNet taxonomy struggles with accurately measuring the distance for verbs, adjectives, and adverbs because these parts of speech have less hierarchical structure and are sometimes unconnected. To overcome this limitation, we measure the distance with the generated encodings discussed in Section~\ref{sec:tax}, enabling a more precise computation of Wu-Palmer similarity.

\subsection{Creating a Challenge Set for  Out-of-Distribution Concepts} \label{subsec:challenge}

\begin{figure}[hbtp]
    \includegraphics[width=0.9\textwidth]{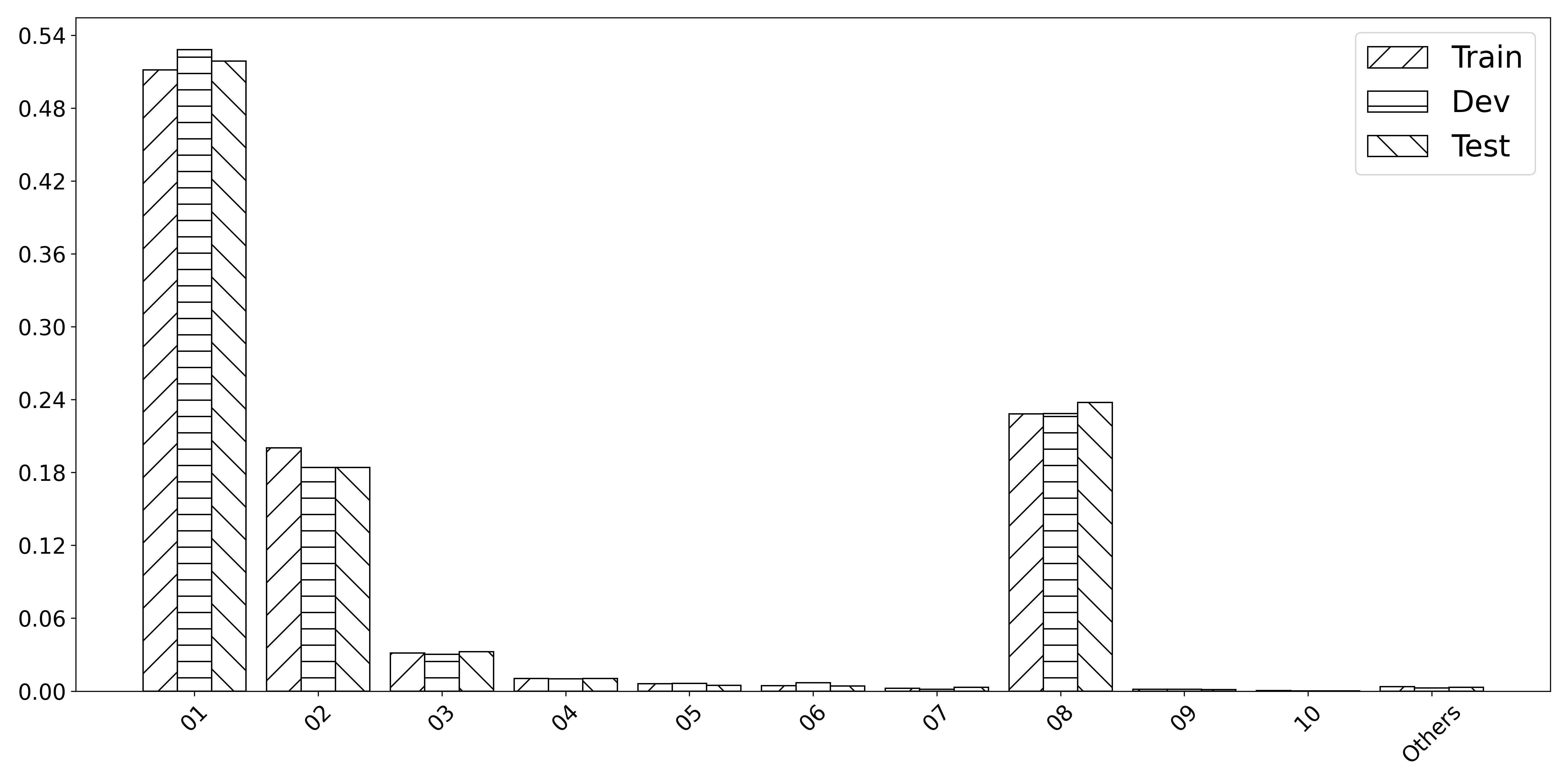}
    \caption{Distribution of word senses in the different data splits of the Parallel Meaning Bank. Note that except sense "01", sense "02" is prominent because every person name incorporates either the \emph{female.n.02} or \emph{male.n.02} and "08" also stands out because every meaning for a tensed clause includes the \emph{time.n.08} concept.}
    \label{fig:sense-number}
\end{figure}

One of our research objectives is to make a model that is able to come up with a reasonable interpretation of a concept that it has never encountered during training.
The Parallel Meaning Bank offers training, development, and test sets, all featuring similar distributions (Figure~\ref{fig:sense-number}). The PMB data indicates that concepts are frequently used with the first word sense. This is not very surprising, because WordNet tends to list the most used senses first for each word.

This poses the following problem to meet our research objective. Say we give the model for semantic parsing a sentence with an unknown word (a word that the model hasn't seen during training). The model will likely transform it into a WordNet concept with sense number $1$, based on the statistics seen on training. Hence, the chance that the model got it correct is very high. But does such a model demonstrate some kind of semantic understanding? Not really---it just produced a pattern that it has seen many, many times during training.

In this case, we need to evaluate capability of dealing with rare or unseen word senses. We do this by creating a challenge test set consisting of more than a hundred English sentences and their gold standard meaning representations, where each sentence contains one or more words (nouns, verbs, or adjectives) that are not part of the training and development set, or are present in the training set with a different sense. To ensure an insightful evaluation, we make certain that the corresponding meaning for these unknown concepts do not correspond to the first sense.
As a source of inspiration we use the glosses and example sentences found in WordNet for a particular word sense, and add enough context to disambiguate the meaning of the word. For instance, in "the moon is waxing", we have the third sense for the verb, resulting in the concept \emph{wax.v.03}, not seen in training (although \emph{wax.v.01} could be part of the training data). For each concept we construct three sentences in which the concept is expressed with enough context for a human to understand the intended meaning.

The entire challenge set comprises  $500$ example sentences paired with their meaning representation in SBN format, containing $430$ unknown nouns, $128$ unknown verbs and $65$ unknown adjectives/adverbs. We verified and manually corrected the annotations when needed to guarantee their gold-standard quality. In  Table~\ref{tab:cha_result} in Section~\ref{sec:experiments}, we showcase some examples along with the predictions of different parsers.

\subsection{Designing a Taxonomy-based Semantic Parsing Architecture}

\begin{figure*}[hbtp]
    \centering
    \includegraphics[width=\textwidth]{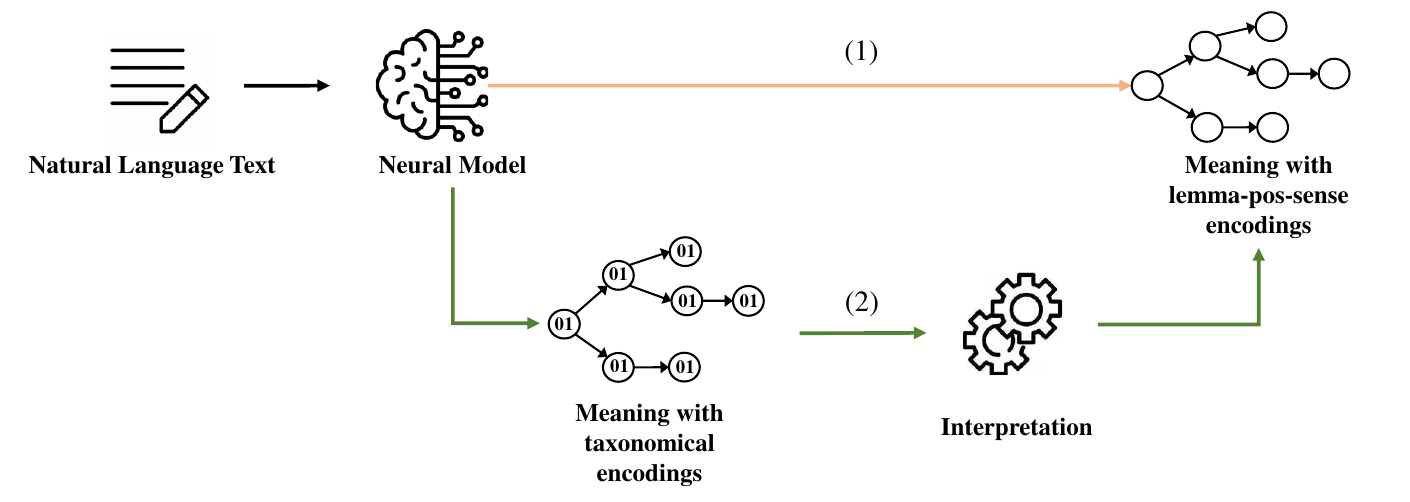}
    \caption{The pipelines for semantic parsing in comparison. Route (1) shows a neural semantic parsing system based on traditional concepts representations. Route (2) illustrates the taxonomy-based parsing system where a mapper interprets the produced symbols.}
    \label{fig:parsing-pipline}
\end{figure*}

Modern semantic parsing predominantly employs sequence-to-sequence models trained with linearized meaning representations \cite{barzdins-gosko-2016-riga, noord-2017-neural, Bevilacqua2021OneST}. 
In our approach, we retain the sequence-to-sequence architecture but adapt the output to represent a linearized graph of semantic representation encoded with taxonomical encodings using the technique proposed by \citet{Bos2023IWCS} as presented in Section~\ref{sec:sbn}.
This output representation needs to undergo a process of interpretation (Figure~\ref{fig:parsing-pipline}). This interpretation is implemented by mapping the taxonomical encodings into concepts and relations of our human-readable dictionary. This dictionary consists of WordNet and the ontology of other symbols including the semantic roles. In other words, what the mapper does is 
translating each encoding into a traditional format found in the standard meaning representations.

Let $\cal{D}$ be a dictionary that maps taxonomical encodings to unambiguous predicate symbols, much like as shown in Tables~\ref{tab:hyponyms}--\ref{tab:seq}.
Let $T_n$ denote a taxonomical sequence meaning representation of length $n$, where $T_n = (t_1, t_2, \ldots, t_n)$. 
Then, the interpretation  of taxonomical encodings is defined as a mapping function $\mathcal{M}$ as follows:

\begin{flalign}
& \mathcal{M}(T_n) = (M(t_1), M(t_2), \ldots, M(t_m), \ldots, M(t_n) ) &  \label{equ:M} \\ 
& M(t_m) = \left\{
\begin{array}{ll}
\mathcal{D}(t_m), & \text{if } t_m \in \text{tax-format} \text{ and } t_m \in \mathcal{D} \\
\mathcal{D}(C(t_m)), & \text{if } t_m \in \text{tax-format} \text{ and } t_m \notin \mathcal{D} \\
t_m, & \text{otherwise} \label{equ:M_detail}
\end{array} 
\right. &
\end{flalign}

The function $\mathcal{M}$ in Equation~\ref{equ:M} denotes the symbolic interpretation process, encapsulating the overall mapping of a meaning representation in taxonomical encodings. The function $M$ in Equation~\ref{equ:M_detail} operates on individual elements of this meaning representation. It distinguishes three cases:
(a) If an element $t_m$ strictly follows the tax code format and is listed in the dictionary, it is directly mapped to the corresponding lemma-pos-sense, role, operator or discourse relation format---for instance, n$1222212113423100000000$ is part of $\cal{D}$, and mapped to hobby.n.03;
(b) If an element is not part of the dictionary but a valid taxonomical encoding, it undergoes computation by $C$, a traversal function that sifts through all tax codes in $\cal{D}$ to identify the encoding closest to the input according to the Wu-Palmer similarity metric---for instance, n$1233111111110000000000$ is not part of $\cal{D}$, but approximated by $C$ to n$1233111111100000000000$ which is part of $D$, and then mapped to a WordNet synset, e.g., wheat\_beer.n.01;\footnote{For WordNet-Identifier encodings, which are represented by numbers, we determine the closest synset just by computing the numerical difference between two identifiers.}
(c) Elements that are not encoded are regarded as literals and left unchanged---for instance, $John$ is kept as it is.

Hence, in our parsing system, taxonomical encodings serve as an intermediate representation, not as the final output. The interpretation component in the pipeline (Figure~\ref{fig:parsing-pipline}) generates  a meaning graph that is readable for humans, encoded with lemma-pos-sense information and the usual labels for roles, operators, and discourse relations format (Table~\ref{tab:roles}). The advantages of taxonomical encodings will be revealed in the following experiment sections.

\section{Experiments and Results}\label{sec:experiments}

We will compare three different representation methods for conceptual predicates:
the standard one based on lemma, part of speech and sense number (LPS, henceforth), one based on rather arbitrary WordNet identifiers (WID), and one based on our novel taxonomical encodings (TAX). The data that we use to run our experiments is
drawn from the Parallel Meaning Bank. For evaluation we employ both the standard test set to assess the overall semantic parsing accuracy (for English and German) as well as the challenge set dedicated to measure the ability to interpret concepts not part of the training data (for English only). 
Furthermore, we probe the models by extracting the sense embeddings to get an idea of they reflect the taxonomical information encoded in WordNet.

\subsection{Data}

For our experiments, we selected the gold-standard English and German data from the Parallel Meaning Bank\footnote{We use the PMB 5.1.0 available at \url{https://pmb.let.rug.nl/releases}. We only use gold data for English and German for our experiments. Although the PMB also offers annotated data for other languages, it is of insufficient quantity for effective training. The PMB also provides silver data (partially annotated and verified by experts), but because word senses are not consistently corrected in this part of the data we will not explore it, although in general adding silver data to the training set has been proved to enhance parsing performance \cite{van-noord-etal-2020-character, poelman-etal-2022-transparent,wang-etal-2023-pre}.} as detailed in Table~\ref{tab:pmb}. English data is divided into training, development, and test sets following an 8:1:1 split ratio, and German data follows 4:3:3 split ratio to make the development and test sufficiently large for evaluation purposes. 

\begin{table}[hbtp]
\caption{Data statistics for two languages in the PMB 5.1.0. Words and Chars represents the average number of words and characters in one sample.}
\label{tab:pmb}
\small
\begin{tabular}{c|ccc|rcc|rcc}
\toprule
 & \multicolumn{3}{c|}{Train} & \multicolumn{3}{c|}{Development} & \multicolumn{3}{c}{Test} \\
\cmidrule(lr){2-4}
\cmidrule(lr){5-7}
\cmidrule(lr){8-10}
 & Samples & Words & Chars & Samples & Words & Chars & Samples & Words & Chars \\ \midrule
English & 9,560 & 5.7 & 29.5 & 1,195 & 5.4 & 29.6 & 1,195 & 5.3 & 29.6 \\
German & 1,256 & 5.1 & 29.9 & 936 & 4.9 & 29.7 & 936 & 4.8 & 29.7 \\ \bottomrule
\end{tabular}
\end{table}

\subsection{Experimental Settings\label{subsec:models}}

In our experiments, we utilized the most frequently used seq2seq architecture, specifically leveraging the transformer-based T5 \cite{Raffel2019ExploringTL} and BART \cite{lewis-etal-2020-bart}, two pre-trained transformer-based architectures. Specifically, we fine-tuned their multilingual variants (because we include both English and German): mT5 \cite{xue-etal-2021-mt5}, byT5 \cite{xue-etal-2022-byt5} and mBART \cite{liu-etal-2020-multilingual-denoising}. mT5 builds upon the T5 model with pre-training on multi-languages corpus. byT5 enhances the multilingual approach through byte-level processing, making it especially effective at handling languages with limited data resources. Meanwhile, mBART also leverages a multilingual corpus for its pre-training phase and adopts a denoising autoencoder strategy. A noteworthy distinction between these models lies in their tokenization approaches: mT5 and mBART utilize sub-word tokenization, while byT5 employs byte-level tokenization. 
In our task, whether employing the lemma-pos-sense notation, WordNet identifiers, or taxonomical encodings, each notation presents a format distinct from the natural languages that were seen in their pre-training corpora. This divergence challenges the tokenization strategies of the models and their proficiency in processing new language with limited data.

We set the learning rate to $10^{-4}$, included a decay rate of $0.5$ and set a patience threshold of $5$ for early stopping. More details are provided in Appendix~\ref{app:exp-setting}. Given the relatively small size of the German dataset for fine-tuning (1,256 instances for training), we initially fine-tune the models on the English data before fine-tuning them with the German data. Each experiment is ran three times to calculate the average and standard deviation, which are detailed in the results tables.\footnote{Code is available at \url{https://github.com/LastDance500/neural-symbolic-parsing}.}

\subsection{Results on Semantic Parsing} 

The results for semantic parsing for each of the three different meaning representations are presented in
Table~\ref{tab:en-result} (English) and Table~\ref{tab:de-result} (German). 
We show the standard (hard) Smatch score (HSm) for exact semantic matching and the soft Smatch score (SSm) for approximate semantic matching. We also include the rate of ill-formed output (IFR), as the seq2seq architectures that we employ do not guarantee well-formed graph meaning representations (any output that is ill-formed is assigned a score of $0$).

\begin{table}[hbtp]
\caption{Semantic parsing results (Hard Smatch, Soft Smatch, Ill-Formed Rate) for \textbf{English} using three different models: Lemma-Pos-Sense, WordNet-IDs, and Taxonomical encodings.}
\label{tab:en-result}
\small \setlength{\tabcolsep}{1.68pt}
\begin{tabular}{c|ccc|ccc|ccc}
\toprule
 & \multicolumn{3}{c|}{LPS} & \multicolumn{3}{c|}{WID} & \multicolumn{3}{c}{TAX} \\
\cmidrule(lr){2-4}
\cmidrule(lr){5-7}
\cmidrule(lr){8-10}
 & HSm & SSm & IFR & HSm & SSm & IFR & HSm & SSm & IFR \\ \midrule
mT5 & 84.2 $\pm 2.6$ & 86.4 $\pm 2.1$  & 6.4  $\pm 0.8$ & 81.1 $\pm 2.5$ & 86.0 $\pm 2.3$ & 4.9  $\pm 0.9$ & 80.1 $\pm 3.2$ & 86.2 $\pm 1.8$ & 3.6 $\pm 0.9$ \\
byT5 & \textbf{87.4} $\pm 1.8$ & 89.4 $\pm 2.3$ & 4.7 $\pm 0.4$ & 86.3 $\pm 1.1$  & 91.2 $\pm 1.0$  & 1.8 $\pm 0.5$  & 86.6 $\pm 2.3$ & \textbf{91.8} $\pm 2.5$ & 2.3 $\pm 0.6$ \\
mBART & 79.5 $\pm 1.2$ & 82.8 $\pm 2.2$  & 3.9  $\pm 0.7$ & 76.4 $\pm 0.9$ & 81.5 $\pm 0.5$ & 3.8 $\pm 0.6$ & 83.0 $\pm 2.6$ & 86.2 $\pm 1.6$ & 3.4 $\pm 0.4$ \\ \bottomrule
\end{tabular}
\end{table}

The standard Smatch scores are a little bit below earlier reported F-scores on PMB data (a Hard Smatch score $94.7$ for English and $92.0$ for German, as reported by \citealp{wang-etal-2023-pre}), but they can be considered decent given that we only train on the gold data part and we don't employ further pre-training on silver and bronze data from the PMB corpus. Besides, our research objective is not to reach the highest performance, but rather compare performance of differently structured predicate symbols in meaning representations. The results for German (Table~\ref{tab:de-result}) are  lower than those for English. We think this is caused by two factors: there is, compared to English, less typographical correspondence between the input words and output meanings, and there is less training data available for German. Nonetheless, the results for German are in line with those for English.

\begin{table}[hbtp]
\caption{Semantic parsing results (Hard Smatch, Soft Smatch, Ill-formed Rate) for \textbf{German} using three different models:  Lemma-Pos-Sense, WordNet-IDs, and taxonomical encodings.}
\label{tab:de-result}
\small \setlength{\tabcolsep}{1.7pt}
\begin{tabular}{c|ccc|ccc|ccc}
\toprule
 & \multicolumn{3}{c|}{LPS} & \multicolumn{3}{c|}{WID} & \multicolumn{3}{c}{TAX} \\
\cmidrule(lr){2-4}
\cmidrule(lr){5-7}
\cmidrule(lr){8-10}
 & HSm & SSm & IFR & HSm & SSm & IFR & HSm & SSm & IFR \\ \midrule
mT5 & 78.6 $\pm 2.2$  & 81.7 $\pm 2.4$ &  \hspace*{-5pt} 8.2 $\pm 1.7$ & 78.5 $\pm 2.3$ & 83.5 $\pm 2.1$ & \hspace*{-5pt} 5.5 $\pm 1.3$  & 77.3 $\pm 2.4$ & 84.5 $\pm 2.5$ & 2.8 $\pm 1.5$ \\
byT5 & \textbf{80.5} $\pm 1.1$ & 82.3 $\pm 1.2$ & 4.5 $\pm 0.6$ & 79.3 $\pm 1.0$ & 86.4 $\pm 1.3$ & 5.8 $\pm 0.8$& 80.1 $\pm 1.2$ & \textbf{88.6} $\pm 1.3$ & 2.2 $\pm 0.4$ \\
\hspace*{-5pt} mBART & 78.3 $\pm 2.3$ & 83.2 $\pm 2.5$ & 1.6 $\pm 0.6$ & 72.5 $\pm 2.4$ & 79.2 $\pm 2.3$  & 4.1 $\pm 0.7$ & 76.3 $\pm 2.1$ & 84.8 $\pm 1.1$  &  \textbf{1.7} $\pm 0.7$ \\ \bottomrule
\end{tabular}
\end{table}

It is  interesting to compare the performance of the three different architectures: mT5, byT5, and mBART. Despite the fact that all models are based on the seq2seq and encoder-decoder framework, they exhibit non-consistent performance on these three representations. This is due to the difference in their pre-training objectives and corpora, activation functions, parameter initialization and other aspects (we kindly refer the reader to the original papers of these models, see Section~\ref{subsec:models}). For instance, for both English and German, mT5, when trained using the WID representation, shows superior Hard Smatch score compared to TAX; in contrast, byT5 and mBART get higher Hard Smatch scores using TAX than using WID (Table~\ref{tab:en-result} and~\ref{tab:de-result}).

The byT5 model achieved the highest Smatch scores among all settings and both languages. We believe that the tokenization strategy of byT5 is the main reason for its good performance. This is in line with the findings of \citet{van-noord-etal-2020-character}, who suggest that DRS parsing benefits from the character-level tokenization. Giving a concrete example, mT5's sub-word tokenizer segments $hobby.n.03$, $101612476$, and $n1222212113423100000000$ into disjointed chunks like $[hobby, ., n, .03]$, $[10, 1612, 476]$, and $[n, 1222, 2121, 1342, 31, 00000000]$, respectively. In contrast, byT5's byte-level tokenizer processes the same inputs into $[h, o, b, b, y, ., n, ., 0, 3]$, $[1, 0, 1, 6, 1, 2, 4, 7, 6]$, and $[n, 1, 2, 2, 2, 1, 2, 1, 1, 3, 4, 2, 3, 1, 0, 0, 0, 0, 0, 0, 0, 0]$, offering a more meaningful and robust segmentation. This distinction is particularly crucial for taxonomical encodings, where each character represents a specific layer within the taxonomy. 

In the rest of this discussion we will focus on byT5 given its superior performance. 
Based on the \textbf{Hard Smatch} scores, LPS  emerged as the top performer, achieving scores of $87.4$ for English and $80.5$ for German. WID and TAX perform similarly: both are slightly lower than LPS. We think the main reason why LPS outperforms TAX and WID with hard smatch is that LPS benefits from (a) generating a lemma by copying character sequences from the text input to the meaning output, and (b) generating the most frequent sense number “01” (see Section 5.2) and thereby producing the correct predicate symbol. Evidences for (a) can be found in Appendix E, where we feed misspelled words to the models revealing copying behaviour for LPS. Evidences for (b) can be found in Appendix F where the majority of sense numbers chosen is “01” for LPS. The predicate symbols for TAX and WID are not based on lemmas and sense numbers (see Table \ref{tab:all_noun}), so they cannot "benefit" from copying lemmas or producing the most frequent sense.

However, the situation changes when we turn to the results of approximate semantic matching, where the TAX-parser demonstrates better performance (91.8 for English and $88.6$ for German). The \textbf{Soft Smatch} score of TAX-parsers improved by a minimum of $3.2$ points over Hard Smatch, reaching a peak increase of $6.2$ for English and $8.5$ for German. Conversely, while the Soft Smatch scores for both LPS and WID saw a modest rise, both the magnitude of their increases and final Soft Smatch scores fell short when compared to the TAX-parser. In this case, the larger increase demonstrates that TAX is doing something interesting that LPS and WID are not capable of. In Section~\ref{sec:conceptresults} and~\ref{sec:probing}, we will perform a deeper analysis of this behaviour.

Considering the ill-formed rate (IFR) we can observe  that there are two main causes for ill-formed outputs. One is that the index points to a non-existent concept, and the other is that the generated graph is cyclic. We found that both the WID-parser and TAX-parser have significantly reduced the frequency of index-related prediction errors, which in case reduce the IFR for both English and German. These errors typically stem from the model's limited understanding of the generated graph structure. The low IFR can be seen as the evidence that proposed uniform representation using encodings enhances the seq2seq model's comprehension of semantic graph structures.

\subsection{Results on Unknown Concept Identification}\label{sec:conceptresults} 

Although the overall results for semantic parsing already favor our newly proposed taxonomical encodings (TAX), we also want to show that TAX is making fewer absurd predictions than its alternatives, LPS and WID. In Section~\ref{subsec:challenge} we presented a challenge test set for semantic parsing that contains out-of-distribution concepts.
There are two ways to look at the results of the three different approaches on this stress test: globally, using Smatch scores; or locally, looking in detail on how the three approaches react on unknown concepts.
The global results in terms of Hard Smatch and Soft Smatch metrics remain in line with the results of the standard tests of the previous section: for ByT5, the LPS-parser scores the highest Hard Smatch ($73.3$) while the TAX-parser scores the highest Soft Smatch ($78.1$).

\begin{table}[h]
\caption{Results on unknown concept identification for nouns, verbs, and adjectives, comparing meaning representations based on the standard lemma-pos-sense notation (LPS),
WordNet identifiers (WID), with taxonomical encodings (TAX).}
\label{tab:challenge-results}
\begin{tabular}{lrccc}
\toprule
Category & \multicolumn{1}{c}{Number} & LPS & WID &  TAX \\
\midrule
noun      &  430   & 0.275  & 0.391 & \textbf{0.421} \\
verb      &  128   & 0.270  & 0.289 & \textbf{0.313} \\
adjective \& adverb &    65   & \textbf{0.447} & 0.410 & 0.432 \\
\bottomrule
\end{tabular}
\end{table}

For a more fine-grained analysis we looked at how the three approaches dealt with the unknown concepts. For each example sentence in the challenge test set 
we identified the unknown concepts and paired them with the prediction of the corresponding concepts in the output meaning representation. This was done via an automatic alignment of concepts followed by human verification and correction when needed. Then we applied the Wu-Palmer similarity to each concept pair (gold vs. predicted).
The results of identifying these unknown concepts are presented in Table~\ref{tab:challenge-results}. 

As Table~\ref{tab:challenge-results} shows, none of the three approaches performs very well.
Recall that achieving a perfect score on this task is highly unlikely --- only by chance a parser might pick the correct word sense. Hence, all three parsers are expected to make mistakes, but  there are important differences in the severity of these mistakes (Table~\ref{tab:cha_result}).

The taxonomical encodings (TAX) show the best performance for unknown noun and verb concepts. The standard notation following the lemma-pos-sense (LPS) convention
yields mediocre results because the parser will in most cases default to the most frequent (first or second) sense following the sense distribution in the training data. The WID (WordNet IDs) parser performs surprisingly well, so the identifiers exhibit some systematic grouping that we are not aware of.
Unknown verb concepts seem harder to predict, perhaps because they show less hierarchical structure in WordNet than nouns. 

The LPS-parser makes the best predictions for adjectives and adverbs.
This can probably be attributed to three factors. Firstly, compared to nouns and verbs, there is no hierarchical structure in WordNet for adjectives and adverbs (see Section~\ref{sec:adj&adv}). Secondly, the Wu-Palmer measure that we use for similarity is not optimal for adjectives and adverbs, as it doesn't take polarity into account in a principled way.\footnote{To give an idea of the complexity of defining similarity of adjectives, consider the comparison of long.a.01 (temporal), long.a.02 (spatial) with short.a.01 (temporal). In a way, long.a.01 and long.a.02 are similar in polarity, but dissimilar in dimension. From a different perspective, long.a.01 and short.a.01 are similar in dimension, but dissimilar in polarity. It is hard to catch this into one single similarity score.}
Thirdly, the training set includes only a modest number of adjectives (2,845) and adverbs (665), limiting the model to effectively learn the taxonomical information inside the encodings. In contrast, with adequate data for nouns (35,836) and verbs (8,620), the model  significantly benefits from taxonomical information.

Table~\ref{tab:cha_result} shows some challenge set examples of unknown concept predictions for the three different parsing models (the results of the entire challenge set are shown in Appendix~\ref{app:all_challenge}).
 As we have seen before, the \textbf{LPS-parser} is extremely good at transforming word forms to a lemma and a high frequency sense number, but this strategy does not fare well on the challenge set. In fact, it makes severe mistakes, predicting \emph{thrush} (the infection) instead of \emph{thrush} (the bird), or \emph{Micky Mantle} (the baseball player) rather than \emph{mantle} (the garment).
However, there were several cases where the LPS-parser had a "lucky strike", when it picked the second sense for a lemma which happened to be correct. A case in point is \emph{Hungarian} (the language), where the LPS-parser 
picked the second sense, perhaps because most languages in WordNet happen to be assigned the second sense (the first sense is usually the inhabitant of a country).

\begin{table}[h]
\setlength{\tabcolsep}{2pt}
\small
\caption{Some instances of the challenge set with words with out-of-vocabulary concepts in bold face, and the concepts predicted by the lemma-pos-sense parser (LPS) and the taxonomical encoding parser (TAX). In brackets the Wu-Palmer similarity score between gold and prediction.}
\label{tab:cha_result}
\begin{tabular}{lcrr}
\toprule
Input Text                          & Gold   & LPS prediction & TAX prediction \\ 
\midrule
Scientist examines the insect's \textbf{antennae}.              & n.03 & antennae.n.01 (0.00) & muscle.n.01 (0.63)\\
... went birdwatching. She saw ... a \textbf{hobby}. & n.03           & hobby.n.01 (0.09) & big\_cat.n.01 (0.67) \\
They played \textbf{Scrabble} in the living room.            & n.02 & scrabble.n.01 (0.11) & chess.n.02 (0.90)\\
The \textbf{thrush}'s song filled the forest with ...  & n.03 & thrush.n.01 (0.17) & pigeon.n.01 (0.79)  \\ 
The soldier was shot in the \textbf{calf}.                & n.02 & calf.n.01 (0.20) & cheek.n.01 (0.58)\\
David was armed with a \textbf{sling}. & n.04 & sling.n.01 (0.20) & gun.n.01 (0.90)\\
Jennifer cooked the bass in a \textbf{steamer}.        & n.02 & steamer.n.01 (0.20) & refrigerator.n.01 (0.43)\\
... mayor proposed extensive \textbf{cuts} in the ... & n.19 & cut.n.01 (0.24) & trade.n.01 (0.52)\\
Tiger Woods \textbf{aced} the 16th hole.                 & v.03 & ace.v.02 (0.25)& dig.v.02 (0.20)\\
... musician was playing a ... \textbf{fugue} ... & n.03 & fugue.n.01 (0.25) & tune.n.01 (0.71)\\
He \textbf{shuffled} the cards.                             &  v.02 & shuffle.v.01 (0.25) & toss.v.03 (0.22)\\
The moon is \textbf{waxing}.                                & v.03 & wax.v.01 (0.25) & wake\_up.v.02 (0.76)\\
The elephant's \textbf{trunk} is an extended nose.           & n.05 & trunk.n.02 (0.26) & ear.n.01 (0.73) \\
The \textbf{stripper} in the club did a strip for us.        & n.03 & stripper.n.01 (0.27) & sailor.n.01 (0.73)\\
She \textbf{dressed} the salad.                              & v.10 & dress.v.01 (0.29) & repair.v.01 (0.25)\\
The woman wore a short black \textbf{mantle}.                & n.08 & mantle.n.02 (0.36) & coat.n.01 (0.86)\\
The athlete had a \textbf{muscular} build.                & a.02 & muscular.a.01 (0.50) & fat.a.01 (0.50)\\
The artist painted with \textbf{vivid} colors.  & a.03 & vivid.a.01 (0.50) & infinite.a.01 (0.59)\\
A tiny \textbf{wren} was hiding in the shrubs.               & n.02 & wren.n.01 (0.54) & oriole.n.01 (0.88)\\
\textbf{Hungarian} is a challenging language ...  & n.02 & hungarian.n.n.02 (1.00) & french.n.02 (0.54)\\
... was playing a ... fugue on the \textbf{grand}.& n.02 & grand.n.02 (1.00) & restaurant.n.01 (0.52)\\
\bottomrule
\end{tabular}
\end{table}

Most interestingly, in analogy with recent approaches to image classification \cite{mukherjee,makingbettermistakes},
the \textbf{TAX-parser} "makes the best mistakes", as it often  predicted concepts similar to the unknowns.
Table~\ref{tab:cha_result} shows some intriguing examples.
For instance, when given the sentence "Jennifer cooked the bass in a steamer", 
it predicts \emph{refrigerator} which is close in meaning to \emph{steamer} (the cooking utensil sense) as they are both appliances.
For the sentence "The soldier was shot in the calf", it predicts
\emph{cheek} (human face) which is close in meaning to \emph{calf} (part of a human leg) as they are both body parts. 
And for the sentence "The woman wore a short black mantle", it predicts
\emph{coat} which is close in meaning to \emph{mantle} (a sleeveless garment) as they are both pieces of clothing. 

In other words, the TAX-parser makes mistakes, but less drastic than the mistakes made by the LPS-parser, because it will attempt to find a concept that is close in meaning (exploiting the contextual understanding of the pre-trained  language model) rather than copying a lemma from the textual input to output meaning, as the LPS-parser seems to be doing.

\subsection{Probing Structural Information in Neural Models} \label{sec:probing}
To check whether our models learn the hierarchical taxonomical information during the training process, we use a probing technique to investigate and understand the internal representations and knowledge encoded within the model. 
Probing is a recent method to validate whether neural models possess certain (structural) properties \cite{ettinger-etal-2016-probing, misra-etal-2023-comps, petersen-potts-2023-lexical}.

In our case, we probe the embeddings of the unknown concepts, given their three corresponding sentences in the challenge set (see Section~\ref{subsec:challenge}). 
Because we want to compare embeddings of different levels of specificity 
to reflect the model's understanding of the WordNet structure, we place each unknown concept into a small hierarchy of four levels based on itself (the most specific level) and its first three hypernyms (increasing in generality on each level). 
To extend coverage, we add the WordNet co-hyponyms to each concept (except for the most general level).

For instance, for the  concept hobby.n.03 the first three hypernyms in WordNet are falcon.n.01, hawk.n.01 and raptor.n.01. 
Then we expand each of these concepts with their co-hyponyms from WordNet. For example, for hobby.n.01 we get the co-hoponyms gyrfalcon.n.01 and kestrel.n.01 among others, and so on.
All these concepts together form what we call a \emph{concept group}.
Each concept group consists of four levels of concepts and is paired with three different sentence templates based on the challenge set. Each sentence template contains exactly one blank in which the lemma corresponding to the concept is filled in (e.g., hobby, kestrel, gyrfalcon, hawk, raptor, etc.).
For our running example, we have the sentence template "Powerful and fast-flying, the \_\_\_ hunts medium-sized birds.", and filling in the blank with the corresponding lemma of the concept group results in a new sentence for each concept. The sentence templates are used for all four levels in the concept group. Table~\ref{tab:probing_examples} shows two concept groups with sentence templates.

\begin{table}[htbp]
\small
\caption{Two examples of concept groups for four levels of specificity. Each group is connected to three sentence templates. Templates are simplified and not all synset instances are shown due to space constraints.}
\label{tab:probing_examples}
\setlength{\tabcolsep}{2pt}
\begin{tabular}{cll} 
\toprule
Level & Concept & Sentence Templates\\
\midrule
0 & drive.n.10, adapter.n.02, airfoil.n.01, ...   & The technician installed the new \_\_\_\_ in \\
1 & device.n.01, ceramic.n.01, connection.n.03, ... & the machine. $|$ She carefully examined the\\
2 & instrumentality.n.03, article.n.02, block.n.01, ... & \_\_\_ for any defects.  $|$ The engineer needed  \\ 
3 & artifact.n.01 &  the specific \_\_\_\_ to complete the project. \\
\midrule
0 & almond.n.02, cherry.n.03, drupelet.n.01, ... & The botanist carefully studied the \_\_\_\_\_ \\
1 & drupe.n.01, achene.n.01, acorn.n.01, ... & under the microscope. $|$ The farmer har-\\
2 & fruit.n.01, agamete.n.01, antheridium.n.01, ... & vested the \_\_\_ from the field. $|$ She placed\\ 
3 & reproductive\_structure.n.01 & the \_\_\_ into the basket during ...\\
\bottomrule
\end{tabular}
\end{table}

This way, for each level in the concept group we obtain around 25 sentences.
We input these sentences to  the model and extract the embeddings of the concepts in the last layer of the model's encoder.
We average the embeddings of all lemmas for each sentence template and do this for each level. This gives us four embeddings for each concept, ranging from specific (e.g., hobby, gyrfalcon, ...) to general (e.g., raptor).

To evaluate the reflection of the hierarchical information captured by the embeddings, we can compare the semantic distances of the embeddings representing the four levels of specificity. The assumption here is that the more generic a concept, the greater the semantic distance to a  specific concept should be. We do this by computing the cosine distance for all combinations of specificity levels (with $n$ levels, these gives us ${n \choose 2}$ distances). 
For the four levels that we have, we obtain six distance pairs, and a total of 15 comparisons to make. Each comparison is evaluated as "satisfied" (following the WordNet hierarchy) or not. 
For instance, the distance of a specific concept (e.g., hobby.n.03) to a slightly more generic concept (e.g., falcon.n.01) should be smaller than the distance of that concept to the most generic concept (e.g., bird.n.01). 

The final score is  computed by the number of satisfied comparisons divided by the total number of comparisons (Table~\ref{tab:prob_result}). We call this metric the
Hierarchy Reflection Score (HRS), and the pseudo code for the metric is shown in Appendix~\ref{app:HRS}. The score (a number between 0 and 1) reflects the hierarchical structure in the model: the higher the score, the closer it follows WordNet's ontology.
We distinguish two variants of HRS: 
\emph{base} and \emph{all}. The base-HRS metric only compares the distances to the most specific concept, whereas all-HRS compares all levels.\footnote{Assuming $d(i,j)$ denotes the distance between embeddings on levels $i$ and $j$, the comparisons for HRS-base  are: $d$(0,1)$<$$d$(0,2), 
                         $d$(0,1)$<$$d$(0,3),
                         $d$(0,2)$<$$d$(0,3) 
and the comparisons for HRS-all are: d(0,1)$<$d(0,2), 
                                          $d$(0,1)$<$$d$(0,3),
                                          $d$(0,2)$<$$d$(0,3),
                                          $d$(1,2)$<$$d$(0,2),
                                          $d$(1,2)$<$$d$(1,3),
                                          $d$(1,3)$<$$d$(2,3),
                                          $d$(2,3)$<$$d$(0,3).}
In our experiment, we compare three different models (LPS-byT5, WID-byT5 and TAX-byT5) and pre-trained sense embeddings, SensEmBERT\footnote{For SensEmBERT, we directly retrieve the embeddings from an existing corpus in \url{https://nlp.uniroma1.it/sensembert/}} by \citet{Scarlini2020SensEmBERTCS} for 130 concepts. 

\begin{table}[htbp]
\caption{Hierarchy Reflection Scores  for the concept embeddings of three fine-tuned byT5 parsers and the sense embeddings of SensEmBERT.}
\label{tab:prob_result}
\begin{tabular}{llcc}
\toprule
Parser & Embedding           & HRS--base   &  HRS--all \\ 
           \midrule
&  SensEmBERT \ \ \ \ \ \ & 0.876  & 0.858 \\
           \midrule
LPS&byT5   & 0.771 & 0.723 \\
WID&byT5   & 0.796 & 0.753 \\
TAX&byT5   & 0.810 & 0.784 \\ \bottomrule
\end{tabular}
\end{table}

Among the models we trained for different semantic representations, the TAX-byT5 model achieved the highest score. The WID-byT5 model delivered a moderate performance, while the LPS-byT5 model had the lowest score. This aligns with the results we observed in the semantic parsing task and unknown concept identification task, where the TAX-byT5 model demonstrated superior structural understanding compared to the other two models. SensEmBERT demonstrates superior performance, but it is unsurprising given that it is specifically trained to adhere to the WordNet hierarchy, compared to the other three models we trained on semantic parsing. 

\begin{figure}[htbp]
  \begin{subfigure}[b]{0.49\textwidth}
    \includegraphics[width=\textwidth]{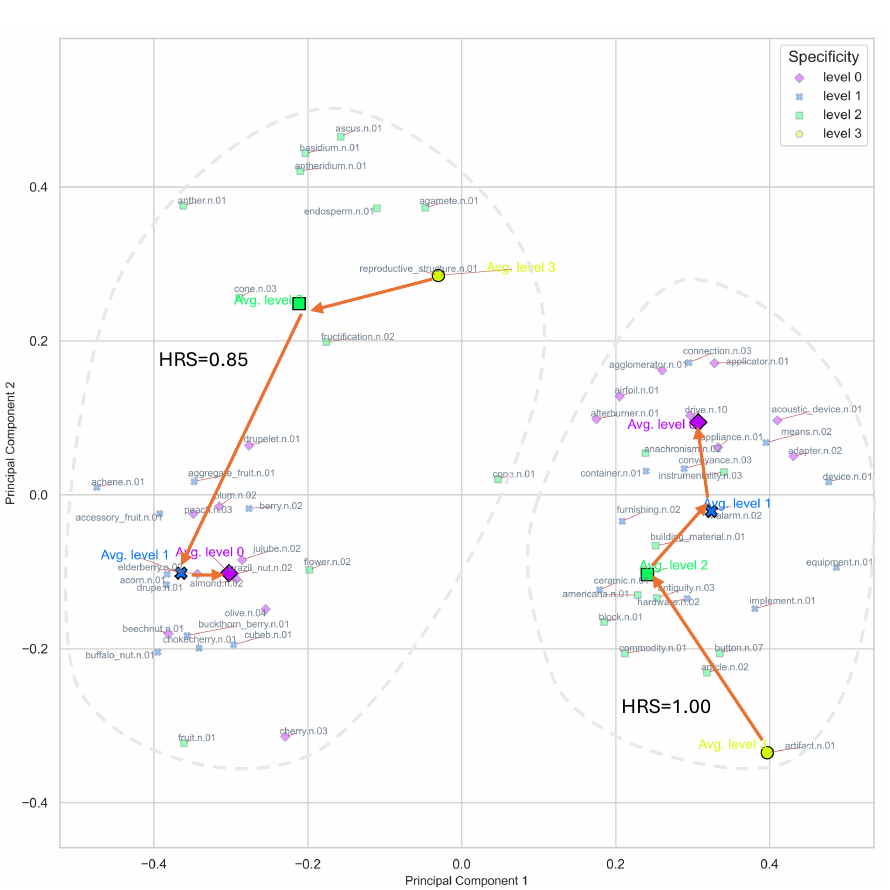}
    \caption{PCA of SensEmBERT Embeddings}
    \label{fig:sen_pca}
  \end{subfigure}
  \hfill %
  \begin{subfigure}[b]{0.49\textwidth}
    \includegraphics[width=\textwidth]{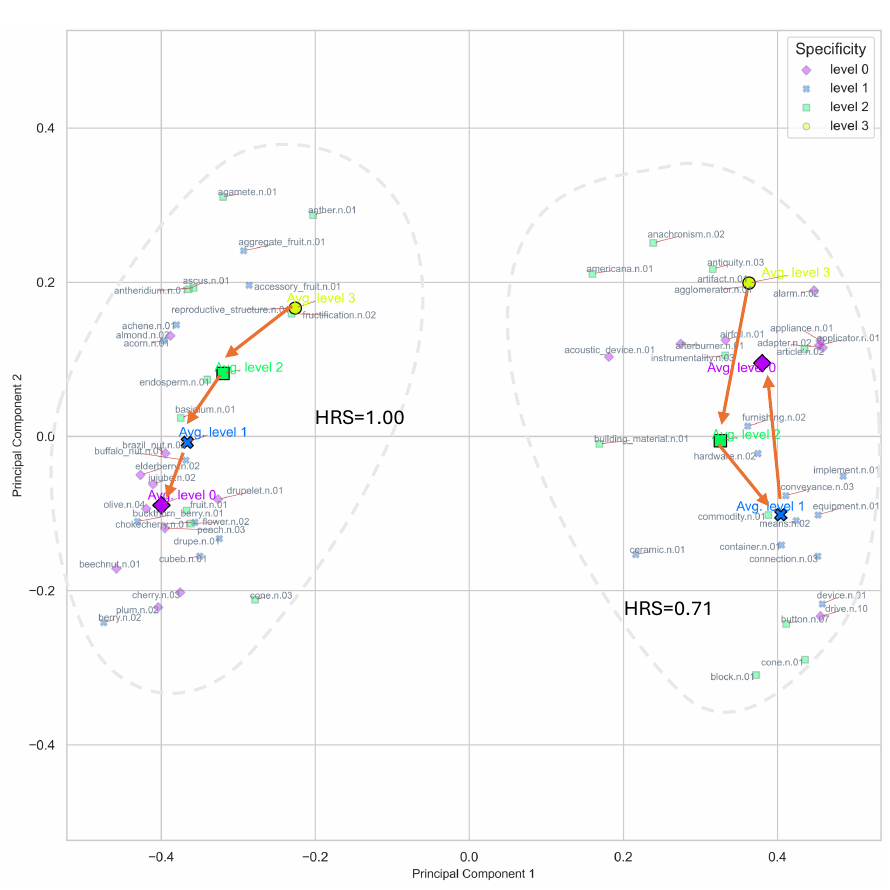}
    \caption{PCA of LPS-byt5 Encoder Embeddings}
    \label{fig:lps_pca}
  \end{subfigure}

  \vspace{1em} %

  \begin{subfigure}[b]{0.49\textwidth}
    \includegraphics[width=\textwidth]{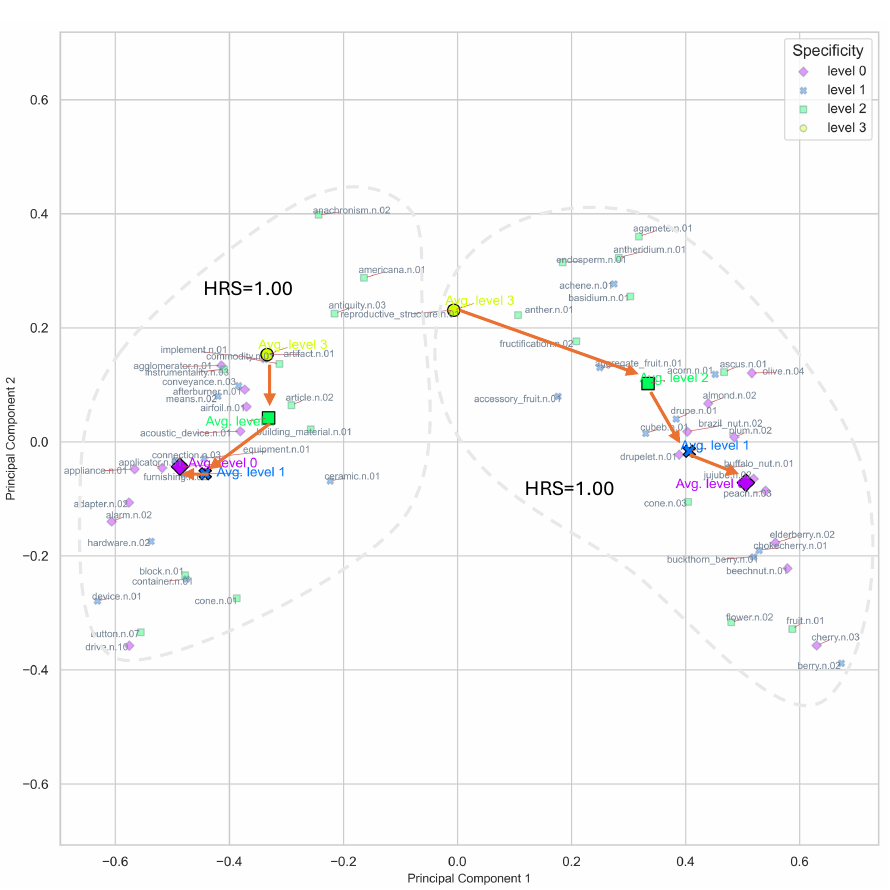}
    \caption{PCA of TAX-byt5 Encoder Embeddings}
    \label{fig:tax_pca}
  \end{subfigure}
  \hfill
  \begin{subfigure}[b]{0.49\textwidth}
    \includegraphics[width=\textwidth]{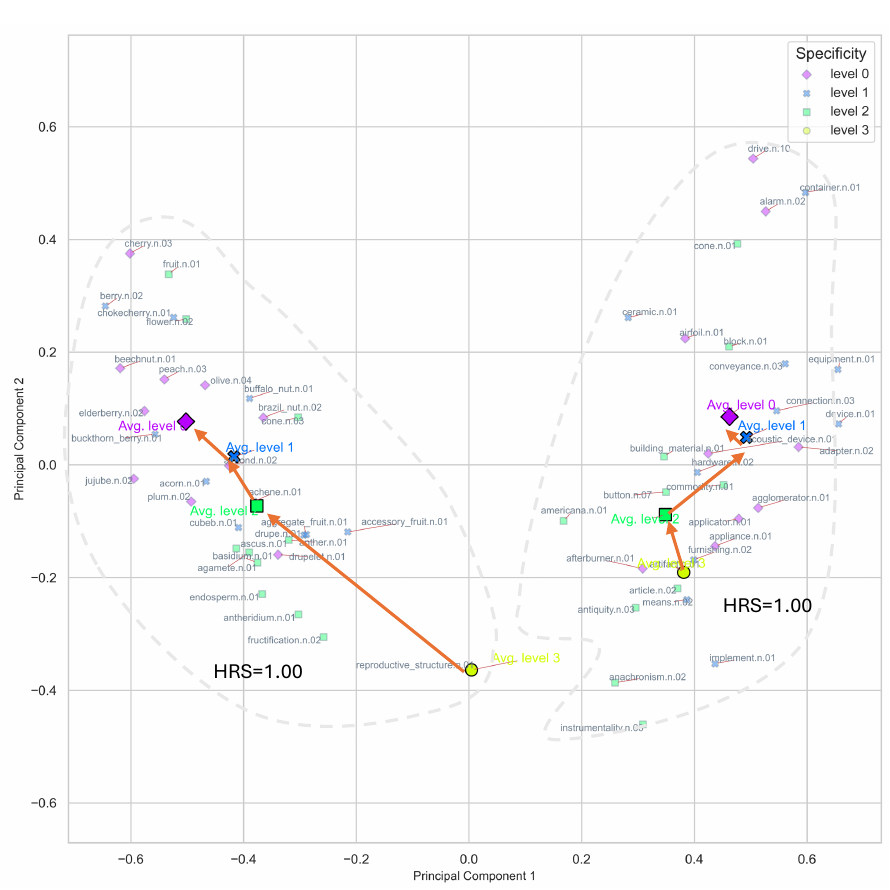}
    \caption{PCA of WID-byt5 Encoder Embeddings}
    \label{fig:wid_pca}
  \end{subfigure}
\caption{PCA analysis of the embeddings for two sets of concepts in Table~\ref{tab:probing_examples}. The orange lines sequentially connect the averaged embeddings of the four specificity levels. The level 0 represents the most specific concepts, and level 3 represents the most general concepts. We add the HRS-all scores for each group of concepts.}
  \label{fig:pca_analyses}
\end{figure}

We can also visualize the results of the probing methods. We follow the method by \citet{lai-nissim-2022-multi} and apply Principal Component Analysis (PCA) to reduce the dimensionality of embeddings. Figure~\ref{fig:pca_analyses} shows the visualisations for the two example sets listed in Table~\ref{tab:probing_examples}.
We use three arrows to connect the averaged embeddings of the four different specificity levels. 
Intuitively, the more the arrows follow a straight line in the same direction, the better they reflect the WordNet hierarchy.
For instance, 
the embeddings of the TAX-parser succeed to reflect the WordNet hierarchy for the conceptual group for driven.n.10, as the arrows in the right of Figure~\ref{fig:tax_pca} form a relatively straight line.
The embeddings of SensEmBERT 
do not entirely reflect the WordNet hierarchy for almond.n.02, as the connected arrows in the left of Figure~\ref{fig:sen_pca} show a slight turn.
The embeddings of the LPS-parser  fail to reflect the WordNet hierarchy driven.n.10, which can be seen by the big turn of the arrows in Figure~\ref{fig:lps_pca}. 
Although Figure~\ref{fig:pca_analyses} only shows the PCA of two instances, it does nicely illustrate the difference in interpretation of hierarchical structure of the models or the lack thereof.

\section{Conclusion and Future Work}

We showed that by taking an existing
lexical ontology, WordNet, we are able to generate hierarchical compositional encodings for predicate symbols for nouns, verbs, adjectives, and adverbs.
We complemented these with encodings for semantic roles, relations, and logical operators. The resulting formal meaning representations contain concept representations that are normalised, abstracting away from specific languages. These extremely rich conceptual representations are still "parseable" for neural models.
For English and German, parsing performance is a little bit lower than the standard lemma-pos-sense notation under Hard Smatch (exact semantic matching), which we attribute to the input-output copying capabilities (translation) of the neural models. However, the advantage of taxonomical encodings is evidenced by a higher Soft Smatch score (approximate semantic matching) and a superior identification of out-of-distribution concepts. Furthermore, the probing results indicate that the models trained with the taxonomical encodings exhibit superior structural understanding capabilities.

We believe that these results are encouraging and a promising way to combine distributional semantics with formal semantics. We hope that the approach presented in this paper is inspirational for future work on neural-symbolic semantic processing. We envision potential both on the symbolic and the neural side.

\smallskip
On the symbolic side, there is a lot of space to further explore the taxonomical encodings. The current encodings are complex, and perhaps there are ways to reduce the number of layers, or different ways of incorporating verbs and adjectives.  Ontologies other than WordNet can be explored, as well as different representations of concepts (perhaps pictograms), and better methods for measuring similarity, in particular that of adjectives and adverbs (especially for the case of antonymy).
Another direction for future work is exploring alternative evaluation metrics to better handle the complexity introduced by the fine-grained similarity evaluation in Soft Smatch. The Soft Smatch method in this article relies on the hill-climbing algorithm of Hard Smatch, which can sometimes result in unwanted matches \cite{opitz-2023-smatch}.\footnote{For instance, if the model predicts (person.n.01, jump.n.01) for the concepts (cat.n.01, laugh.n.01), hill-climbing may match cat.n.01 with jump.n.01 and laugh.n.01 with person.n.01 because these matches score higher than cat.n.01 with person.n.01 and laugh.n.01 with jump.n.01, leading to some spurious scores.}

On the neural side, there are a lot of potential areas worth exploring that have fallen outside the scope of this article. Firstly, incorporating sense embeddings is promising, but integrating them into a neural semantic parser that produces complete meaning representations is challenging.\footnote{Adding pre-trained sense embeddings to the tokenizer could enhance its understanding of WordNet senses, but there are two main drawbacks: (1) compatibility issues due to independently trained embeddings with mismatched dimensions (e.g., 300 for AutoExtend and 2,048 for SensEmBERT, versus 768 for T5-base and 1,024 for T5-large); (2) a significant increase in the tokenizer's dictionary size, given the PMB corpus has over 10,000 senses and WordNet contains more than 100,000 senses.}
Another interesting area of research is to investigate modifications of the loss function aimed at enhancing the model's understanding of taxonomical information in the encodings, where weights are assigned to characters based on their positions.
Another direction to consider is using large language models such as Phi3, Mistral, LlaMa, and GPT-4 for semantic parsing, as they are known to have strong language modeling capabilities. 
However, their architectures (decoder-only) and corresponding experimental settings strongly differ from those of our models. Some pilot experiments that we ran indicate that their performance, whether using standard representations or taxonomical encodings, is by far inferior to our models. An investigation on why large language models perform so poorly on semantic parsing goes beyond the objectives of this article but is perhaps an exciting topic for future resarch.

\section*{Acknowledgements}

We wish to thank the three anonymous reviewers for their helpful suggestions that greatly improved this article. We would also like to  express our gratitude to Huiyuan Lai, Gertjan van Noord and Juri Opitz for their valuable feedback on earlier versions of our work.

\clearpage
\appendix
\appendixsection{Meaning representations in LPS, WID and TAX} \label{app:graphformat}

We show  here the three different types of meaning representations used in our experiments. We do this for the text “John, a keen birdwatcher, was delighted to see a hobby.” in graph format for readability. In the experiments we use the sequence notation.

\begin{figure}[htbp]
    \centering
    \includegraphics[width=\textwidth]{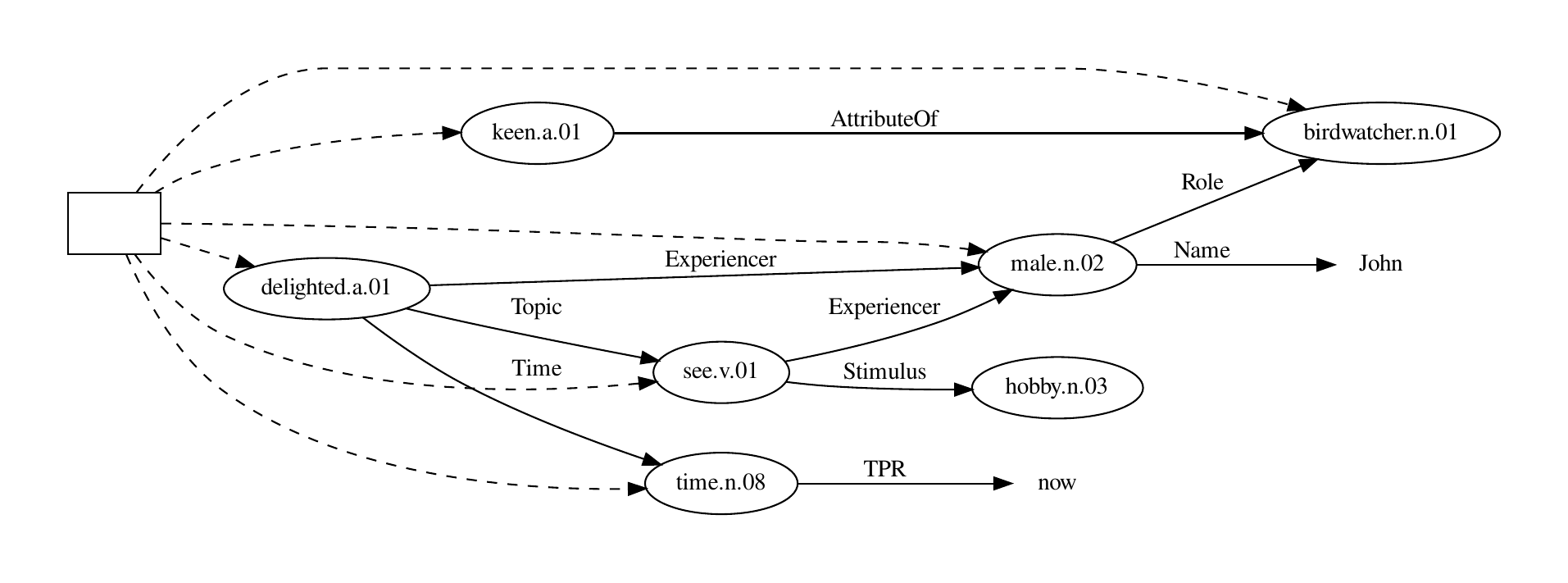}\vspace*{-20pt}
    \caption{Meaning graph using the lemma-pos-sense (LPS) notation to encode concepts.}
    \label{fig:lps-graph}
\end{figure}

\begin{figure}[htbp]
    \centering
    \includegraphics[width=\textwidth]{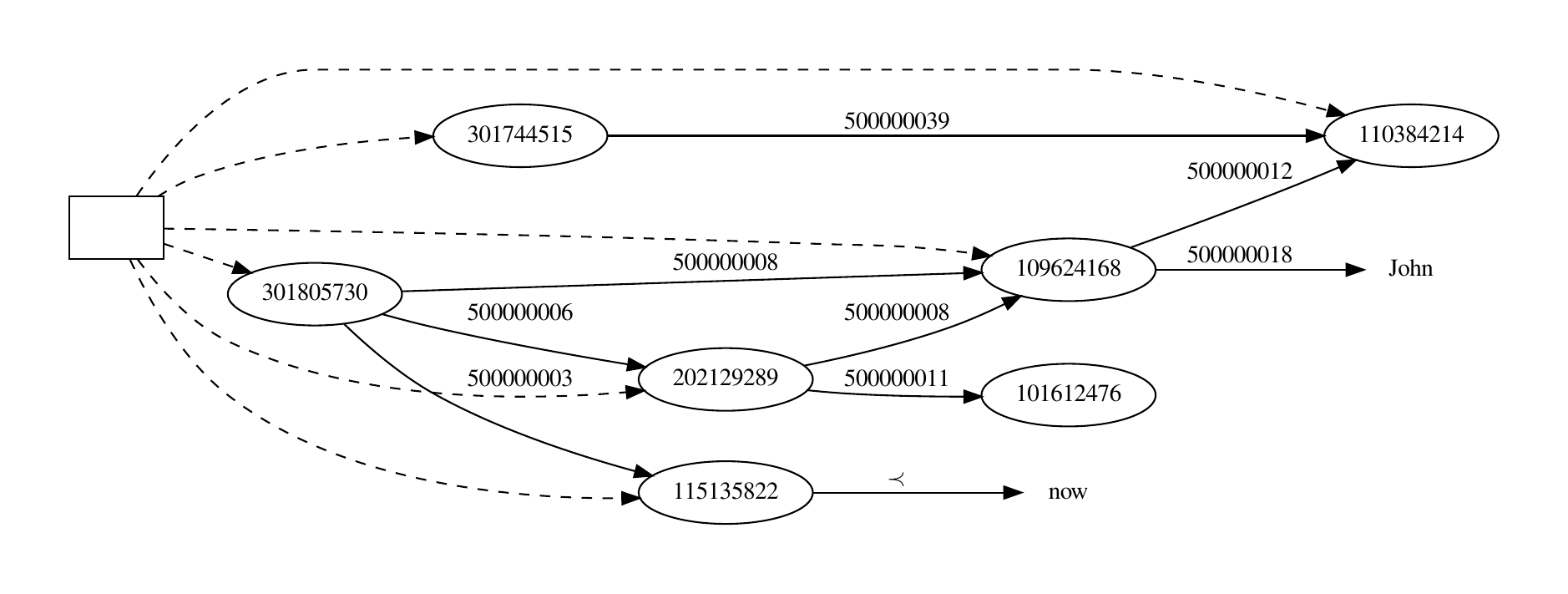}\vspace*{-20pt}
    \caption{Meaning graph based on unique identifiers (WID). The identifiers for synsets are taken from WordNet. The identifiers for roles are assigned by us.}
    \label{fig:wid-graph}
\end{figure}

\begin{figure}[htbp]
    \centering
    \includegraphics[width=\textwidth]{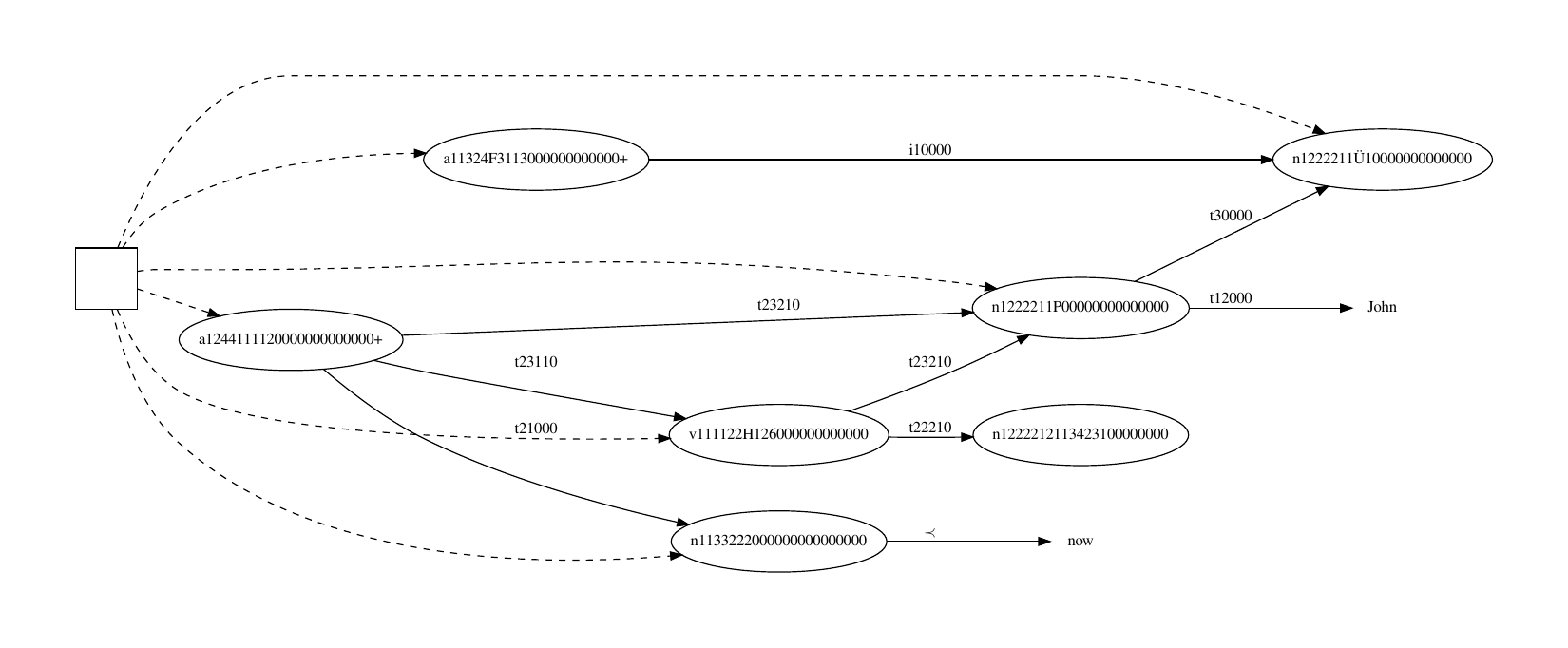}\vspace*{-20pt}
    \caption{Meaning graph with taxonomical encodings for concepts and roles (TAX).}
    \label{fig:tax-code-graph}
\end{figure}

\clearpage

\appendixsection{Taxonomy of Roles used in the Parallel Meaning Bank \label{app:roles}}

Figure~\ref{tab:roles} shows the hierarchy of roles and relations as used in the Parallel Meaning Bank. This hierarchy is an extension of the one proposed for VerbNet \cite{Bonial-2011-verbnet}.

\begin{figure}[hbtp]
    \centering
    \includegraphics[width=\textwidth]{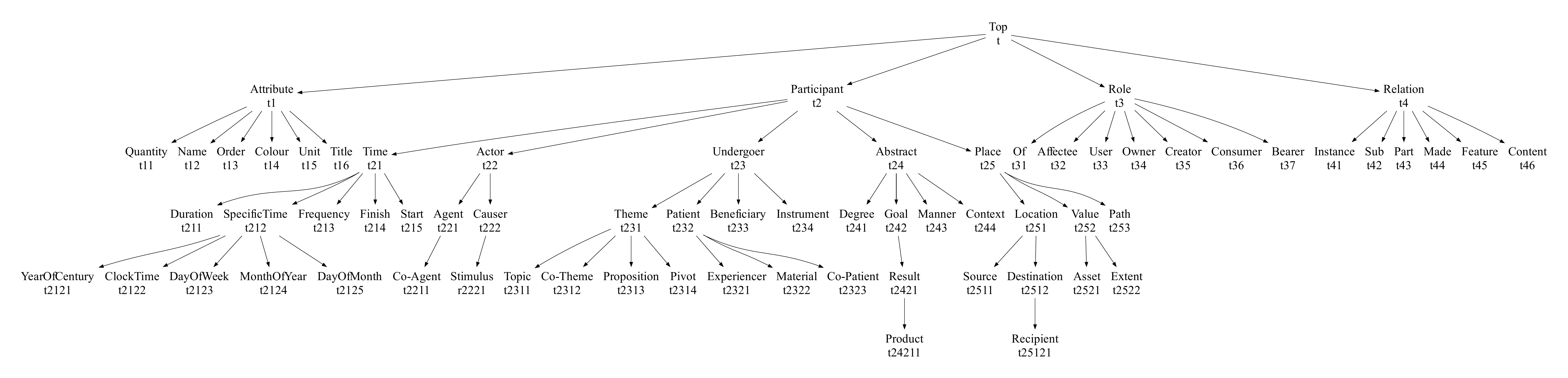}
    \caption{Complete hierarchy of semantic roles used in the semantic parsing experiments. Each role name is shown with the taxonomical encoding.}
    \label{fig:taxonomy-role}
\end{figure}

\clearpage
\appendixsection{Taxonomy encodings of Operators and Discourse Relations.} \label{app:operators}

The identifier and character encoding of operators and discourse relations in Table~\ref{tab:roles} and Table~\ref{tab:discourserelations}.

\begin{table}[hbtp]
\caption{Taxonomy encodings of Operators. Differently from the previous table, the id mentioned here is not the WordNet identifier, but rather one that we have assigned manually.}
\label{tab:roles}
\small
    \begin{tabular}{cccl}
\toprule
    \textbf{Operator} & \textbf{Identifier} & \textbf{Char code} & \textbf{Meaning}     \\ 
    \midrule
    TPR & 700000001 & $\prec$      & temporal precedes (before)\\
    TSU & 700000002 & $\succ$      & temporal succeeds (after)\\
    TIN & 700000003 & $\sqsubset$  & temporal inclusion\\
    TCT & 700000004 & $\sqsupset$  & temporal contains\\
    TAB & 700000005 & $\bowtie$    & temporal abut \\
    LES & 700000006 & $<$          & less than\\
    LEQ & 700000007 & $\leq$       & less or equal than\\
    TOP & 700000008 & $\top$       & not more than\\
    MOR & 700000010 & $>$         & greater than\\
    EQU & 700000011 & $=$         & equal\\
    ANA & 700000012 & $\equiv$    & anaphoric link\\
    APX & 700000013 & $\approx$   & approximately equal\\
    NEQ & 700000014 & $\neq$      & not equal\\
    SXP & 700000015 & $\gg$       & spatially behind\\
    SXN & 700000016 & $\ll$       & spatially before\\
    SZN & 700000017 & $\veebar$   & spatially under\\
    SZP & 700000018 & $\curlyvee$ & spatially above\\
\bottomrule 
    \end{tabular} 
\end{table}

\begin{table}[hbtp]
\caption{Taxonomy encodings of Discourse Relations.}
\label{tab:discourserelations}
\small
\begin{tabular}{ccc}
\toprule
    \textbf{Relation} & \textbf{Identifier} & \textbf{Char code}    \\ 
    \midrule
    ALTERNATION & 600000001 & $\lor$ \\
    ATTRIBUTION & 600000002 & $@$ \\
    CONDITION & 600000003 & $\rightarrow$ \\
    CONSEQUENCE & 600000004 & $\Rightarrow$ \\
    CONTINUATION & 600000005 & $\leftrightarrow$ \\
    CONTRAST & 600000006 & $/$ \\
    EXPLANATION & 600000007 & $\infty$ \\
    NECESSITY & 600000008 & $\square$ \\
    NEGATION & 600000009 & $\neg$ \\
    POSSIBILITY & 600000010 & $\diamond$ \\
    PRECONDITION & 600000011 & $\leftarrow$ \\
    RESULT & 600000012 & $\sum$ \\
    SOURCE & 600000013 & $\hookleftarrow$ \\
    CONJUNCTION & 600000014 & $\land$ \\
    ELABORATION & 600000015 & $\supset$ \\
    COMMENTARY & 600000016 & $\dag$ \\
\bottomrule
\end{tabular}
\end{table}

\clearpage
\appendixsection{Experimental Settings\label{app:exp-setting}}

To facilitate reproduction, we detail the important hyperparameters employed. For the WID and TAX encodings, we adopt a larger patience and decay rate, allowing ample time for convergence. This decision stems from experimental observations indicating that these novel (to the models) representations exhibit a slower convergence rate compared to LPS. We test using hierarchical loss, which give higher weight to the characters on the left within each word, but the initial experiments didn't show any improvements.

\begin{table}[hbtp]
\caption{Hyperparameters for training different representations.}
\begin{tabular}{ccccccc}
\toprule
 & Learning Rate & Epoch & Patience & Decay Rate & Optimizer & Loss Function \\ \midrule
LPS & 1e-4 & 50 & 10 & 0.1 & AdamW & Cross Entropy \\
WID & 1e-4 & 50 & \textbf{10} & \textbf{0.5} & AdamW & Cross Entropy \\
TAX & 1e-4 & 50 & \textbf{10} & \textbf{0.5} & AdamW & Cross Entropy \\ \bottomrule
\end{tabular}
\end{table}

\appendixsection{Interpreting Misspellings \label{app:misspelling}}

In order to assess the ability of models to deal with misspellings
we created a test suite of English sentences paired with meaning representations where each sentence contained a commonly misspelled content word, i.e., \emph{acceptible}, \emph{humourous}, \emph{enterpreneur}.
The results, shown in Table~\ref{tab:misspelling-results}, demonstrate that the traditional lemma-pos-sense notation fails to identify wrongly spelled content words caused by its tendency to copy character sequences from input words to output lemmas.

\begin{table}[hbtp]
\caption{Results on misspelled content words by computing concept identification scores for the three different models: lemma-pos-sense (LPS), WordNet identifiers (WID), and taxonomical encodings (TAX).}
\label{tab:misspelling-results}
\begin{tabular}{lrccc}
\toprule
Category & \multicolumn{1}{c}{Number} & LPS & WID &  TAX \\
\midrule
noun & 16  & 0.105  & 0.441  & \textbf{0.563} \\
verb & 10   & 0.149  & \textbf{0.648} & 0.593 \\
adjective      &  16   & 0.000  & 0.415 & \textbf{0.510} \\
\bottomrule
\end{tabular}
\end{table}

\clearpage
\appendixsection{Concept Identification Results \label{app:all_challenge}}

This appendix shows predictions of the three semantic parsers (LPS, WID, TAX) on the challenge set for nouns (Table 1), verbs (Table 2) and modifiers (Table 3). The challenge set comprises several sentences for the out-of-vocabulary concepts. For reasons of space, only one prediction for each concept is shown. The complete predictions are available on our GitHub site \footnote{\url{https://github.com/LastDance500/neural-symbolic-parsing}}.

\begin{table}[hbtp]
\setlength{\tabcolsep}{9pt}
\caption{Instances of the challenge set with nouns with out-of-vocabulary concepts. In brackets the Wu-Palmer similarity score between gold and prediction.}
\label{tab:all_noun}
\scriptsize

\begin{tabular}{rrrr}

\toprule
Gold  & LPS   & WID & TAX \\ 
\midrule
extract.n.02 & extract.n.01 (0.24) & speech.n.01 (0.35) & history.n.02 (0.47)\\
cruiser.n.03 & cruiser.n.01 (0.36) & dancer.n.01 (0.40) & ship.n.01 (0.85)\\
warbler.n.02 & warbler.n.02 (1.00) & fictional\_animal.n.01 (0.70) & tower.n.01 (0.45)\\
rag.n.03 & rag.n.01 (0.14) & song.n.01 (0.75) & pot.n.01 (0.20)\\
harrier.n.03 & harrier.n.01 (0.46) & tiger.n.02 (0.67) & shed.n.01 (0.42)\\
hobby.n.03 & hobby.n.01 (0.11) & luggage.n.01 (0.40) & lobby.n.01 (0.40)\\
stool.n.02 & stool.n.01 (0.59) & stadium.n.01 (0.35) & tooth.n.01 (0.30)\\
eagle.n.02 & eagle.n.01 (0.19) & eagle.n.01 (0.19) & eagle.n.01 (0.19)\\
wallflower.n.03 & wallflower.n.01 (0.52) & vegetarian.n.01 (0.70) & comedian.n.01 (0.73)\\
beetle.n.02 & beelte.n.02 (0.00) & bed.n.01 (0.61) & shed.n.01 (0.55)\\
stake.n.05 & stake.n.01 (0.13) & storage\_space.n.01 (0.55) & hull.n.06 (0.57)\\
hungarian.n.02 & hungarian.n.02 (1.00) & german.n.02 (0.57) & french.n.01 (0.54)\\
pen.n.02 & pen.n.01 (0.60) & pen.n.01 (0.60) & pen.n.01 (0.60)\\
pen.n.05 & pen.n.01 (0.42) & pen.n.01 (0.42) & pen.n.01 (0.42)\\
gondola.n.02 & gondola.n.01 (0.57) & gun.n.01 (0.58) & bottle.n.01 (0.61)\\
bug.n.03 & bug.n.02 (0.19) & coop.n.02 (0.52) & disease.n.01 (0.17)\\
investigation.n.02 & investigation.n.01 (0.33) & practice.n.04 (0.78) & wrongdoing.n.02 (0.82)\\
thrush.n.03 & thrush.n.01 (0.17) & lemur.n.01 (0.71) & pigeon.n.01 (0.79)\\
song.n.04 & song.n.01 (0.53) & song.n.01 (0.53) & song.n.01 (0.53)\\
admiral.n.02 & admiral.n.01 (0.48) & aunt.n.01 (0.54) & crocodilian\_reptile.n.01 (0.57)\\
flower.n.02 & flower.n.01 (0.45) & flower.n.01 (0.45) & flower.n.01 (0.45)\\
bloom.n.02 & bloom.n.01 (0.33) & blood.n.01 (0.33) & blood.n.01 (0.33)\\
wren.n.02 & wren.n.01 (0.57) & grass.n.01 (0.56) & oriole.n.01 (0.88)\\
bed.n.03 & ocean\_bed.n.01 (0.00) & picture.n.02 (0.47) & beach.n.01 (0.77)\\
impression.n.04 & impression.n.01 (0.38) & tear.n.01 (0.38) & smell.n.01 (0.33)\\
tripper.n.04 & tripper.n.01 (0.40) & tv\_set.n.01 (0.61) & elevator.n.01 (0.76)\\
reel.n.05 & reel.n.02 (0.12) & ranch.n.01 (0.17) & bike.n.02 (0.16)\\
course.n.07 & course.n.01 (0.12) & candy.n.01 (0.78) & cup.n.02 (0.27)\\
mantle.n.08 & mantle.n.02 (0.00) &  (0.00) & coat.n.01 (0.86)\\
joint.n.06 & joint.n.01 (0.32) & jet.n.01 (0.23) & joint.n.01 (0.32)\\
net.n.05 & net.n.02 (0.70) & napkin.n.01 (0.55) & net.n.02 (0.70)\\
rally.n.05 & rally.n.02 (0.47) & initiation.n.01 (0.59) & race.n.02 (0.62)\\
adder.n.03 & adder.n.01 (0.38) & back\_door.n.03 (0.37) & astronaut.n.01 (0.56)\\
key.n.04 & key.n.03 (0.13) & key.n.01 (0.22) & key.n.01 (0.22)\\
harrier.n.02 & hard\_coat.n.01 (0.00) & dog.n.01 (0.91) & joiner.n.01 (0.47)\\
drive.n.10 & drive.n.01 (0.12) & dress.n.01 (0.63) & engine.n.01 (0.80)\\
fugue.n.03 & fugue.n.01 (0.29) & music.n.01 (0.57) & tune.n.01 (0.71)\\
grand.n.02 & grand.n.02 (1.00) & guitar.n.01 (0.78) & restaurant.n.01 (0.52)\\
application.n.04 & application.n.01 (0.30) & comic\_book.n.01 (0.17) & application\_form.n.01 (0.50)\\
bag.n.03 & bag.n.01 (0.70) & bag.n.01 (0.70) & bag.n.01 (0.70)\\
cover.n.09 & cover.n.01 (0.62) & schoolroom.n.01 (0.57) & mask.n.04 (0.60)\\
pain.n.04 & pain.n.01 (0.20) & pain.n.01 (0.20) & pain.n.01 (0.20)\\
stripper.n.03 & stripper.n.01 (0.29) & wizard.n.02 (0.76) & sailor.n.01 (0.73)\\
strip.n.06 & strip.n.02 (0.21) & trip.n.01 (0.50) & trip.n.01 (0.50)\\
substance.n.04 & substance.n.01 (0.67) & object.n.04 (0.36) & drug.n.01 (0.60)\\
ray.n.07 & ray.n.01 (0.21) & hedgehog.n.02 (0.69) & sand.n.01 (0.25)\\
increase.n.05 & increase.n.01 (0.15) & eye\_blink.n.01 (0.21) & rate.n.02 (0.30)\\
cut.n.19 & cut.n.01 (0.27) & art.n.02 (0.60) & trade.n.01 (0.52)\\
antenna.n.03 & antennae.n.01 (0.00) & alarm.n.04 (0.20) & muscle.n.01 (0.63)\\
entrance.n.03 & entrance.n.02 (0.29) & laugh.n.01 (0.38) & landing.n.04 (0.89)\\
\bottomrule
\end{tabular}
\end{table}
\begin{table}[hbtp]
\setlength{\tabcolsep}{9pt}
\label{tab:all_result2}
\scriptsize
\begin{tabular}{rrrr}
\toprule
Gold  & LPS   & WID & TAX \\ 
\midrule
operation.n.05 & operation.n.01 (0.31) & war.n.01 (0.71) & job.n.02 (0.78)\\
service.n.15 & service.n.01 (0.74) & sewing\_machine.n.01 (0.18) & service.n.01 (0.74)\\
whisker.n.02 & whisker.n.01 (0.10) & cat.n.01 (0.25) & mouse.n.01 (0.26)\\
attack.n.07 & attack.n.03 (0.21) & cold.n.01 (0.27) & attack.n.07 (1.00)\\
appearance.n.04 & appearance.n.01 (0.43) & negotiation.n.01 (0.38) & athletic\_game.n.01 (0.44)\\
sub.n.02 & sub.n.01 (0.29) & stuff.n.02 (0.40) & dagger.n.01 (0.52)\\
dock.n.03 & dock.n.01 (0.44) & dog.n.01 (0.40) & dog.n.01 (0.40)\\
touch.n.10 & touch.n.01 (0.47) & view.n.02 (0.53) & improvement.n.01 (0.44)\\
weight.n.07 & weight.n.01 (0.43) & kilo.n.01 (0.75) & value.n.02 (0.43)\\
pan.n.03 & scale\_pan.n.01 (0.00) & sweater.n.01 (0.63) & tumbler.n.02 (0.84)\\
labor.n.02 & labor.n.02 (1.00) & behavior.n.01 (0.82) & job.n.01 (0.82)\\
unit.n.03 & offensive\_unit.n.01 (0.00) & college.n.02 (0.75) & school.n.01 (0.75)\\
period.n.07 & period.n.01 (0.33) & time.n.08 (0.33) & time.n.03 (0.35)\\
top.n.10 & top.n.01 (0.47) & toe.n.01 (0.30) & roof.n.01 (0.74)\\
top.n.09 & top.n.02 (0.42) & top.n.02 (0.42) & frontier.n.02 (0.47)\\
carton.n.02 & carton.n.01 (0.13) & calculator.n.02 (0.70) & ax.n.01 (0.61)\\
trunk.n.05 & trunk.n.02 (0.26) & hat.n.01 (0.27) & ear.n.01 (0.73)\\
organ.n.03 & organ\_onstage.n.01 (0.00) & company.n.01 (0.21) & piano.n.01 (0.82)\\
cape.n.02 & cape.n.02 (1.00) & calculator.n.02 (0.55) & wash.n.07 (0.82)\\
song.n.05 & song.n.01 (0.33) & college.n.02 (0.32) &  (0.27)\\
heat.n.06 & heat.n.02 (0.38) & hour.n.01 (0.40) & heat.n.02 (0.38)\\
mouth.n.04 & mouth.n.01 (0.35) & middle.n.01 (0.57) & frontier.n.02 (0.53)\\
calf.n.02 & calf.n.01 (0.42) & calculator.n.02 (0.29) & cheek.n.01 (0.58)\\
chemistry.n.03 & chemistry.n.01 (0.35) & chemistry.n.01 (0.35) & natural\_science.n.01 (0.38)\\
crown.n.07 & crown.n.01 (0.14) & bus\_stop.n.01 (0.59) & haunt.n.01 (0.62)\\
mole.n.03 & mole.n.02 (0.22) & bread.n.01 (0.40) & cup.n.01 (0.26)\\
almond.n.02 & almond.n.02 (1.00) & sugar.n.01 (0.24) & entity.n.01 (0.25)\\
bass.n.04 & bass.n.02 (0.12) & pretzel.n.01 (0.63) & sandglass.n.01 (0.29)\\
steamer.n.02 & steamer.n.01 (0.24) & spoon.n.01 (0.52) & refrigerator (0.43)\\
lock.n.02 & lock.n.01 (0.48) & shit.n.01 (0.30) & screw.n.04 (0.48)\\
ace.n.06 & ace.n.08 (0.00) & extraterrestrial.n.01 (0.32) &  (0.00)\\
slide.n.03 & slide.n.03 (1.00) & soccer\_ball.n.01 (0.20) & sunglasses.n.01 (0.19)\\
slip.n.11 & slip.n.01 (0.11) & sock.n.01 (0.73) & wash.n.07 (0.86)\\
scrabble.n.02 & scrabble.n.01 (0.11) & rugby.n.01 (0.56) & chess.n.01 (0.90)\\
decoy.n.02 & decoy.n.01 (0.52) & fly.n.01 (0.45) & mosquito.n.01 (0.45)\\
jay.n.02 & jay.n.01 (0.50) & hedgehog.n.02 (0.69) & dolphin.n.02 (0.62)\\
hole.n.03 & hole.n.02 (0.29) & hole.n.02 (0.29) & shore.n.01 (0.33)\\
hawker.n.02 & hawker.n.01 (0.55) & hunter.n.01 (0.96) & guest.n.01 (0.70)\\
merlin.n.02 & merlin.n.01 (0.11) & match.n.01 (0.40) & bat.n.01 (0.71)\\
rocket.n.03 & rocket.n.01 (0.48) & rayon.n.01 (0.53) & spoon.n.01 (0.45)\\
move.n.05 & move.n.01 (0.63) & movie.n.01 (0.59) & assignment.n.05 (0.74)\\
barrel.n.02 & barrel.n.02 (1.00) & basket.n.01 (0.84) & balcony.n.02 (0.67)\\
function.n.07 & function.n.01 (0.38) & baseball\_club.n.01 (0.30) & job.n.02 (0.30)\\
string.n.05 & string.n.01 (0.22) & page.n.01 (0.21) & lock.n.01 (0.20)\\
green.n.06 & green.n.02 (0.53) & tomb.n.01 (0.56) & grey.n.05 (0.21)\\
surge.n.03 & surge.n.01 (0.15) & person.n.01 (0.24) & person.n.01 (0.24)\\
wave.n.06 & wave.n.01 (0.21) & quantity.n.01 (0.27) & marker.n.02 (0.22)\\
sling.n.04 & sling.n.02 (0.59) & soccer\_ball.n.01 (0.64) & gun.n.01 (0.90)\\
sling.n.05 & sling.n.01 (0.20) & sock.n.01 (0.64) & canopy.n.03 (0.67)\\
china.n.02 & china.n.02 (1.00) & continent.n.01 (0.42) & orange\_juice.n.01 (0.29)\\
slug.n.07 & slug.n.01 (0.32) & goose.n.01 (0.59) & mosquito.n.01 (0.72)\\
growth.n.04 & growth.n.01 (0.29) & fruit.n.01 (0.24) & flower.n.01 (0.21)\\
bullfinch.n.02 & bullfinch.n.01 (0.57) & metatherian.n.01 (0.74) & chicken.n.02 (0.76)\\
\bottomrule
\end{tabular}
\end{table}

\begin{table}[hbtp]
\caption{Instances of the challenge set with verbs with out-of-vocabulary concepts.}

\setlength{\tabcolsep}{9pt}
\label{tab:all_verb}
\scriptsize
\begin{tabular}{rrrr}
\toprule
Gold  & LPS   & WID & TAX \\ 
\midrule
drive.v.08 & drive.v.01 (0.40) & drive.v.01 (0.40) & drive.v.01 (0.40)\\
house.v.02 & house.v.01 (0.00) &  (0.00) &  (0.00)\\
run.v.22 & run.v.01 (0.15) & run.v.01 (0.15) & run.v.01 (0.15)\\
release.v.05 & release.v.02 (0.27) & step\_out.v.01 (0.20) & throw.v.03 (0.19)\\
serve.v.15 & serve.v.07 (0.25) & serve.v.06 (0.50) & give.v.24 (0.50)\\
run.v.19 & run.v.07 (0.24) & work.v.04 (0.24) & pass.v.14 (0.18)\\
recognize.v.08 & recognize.v.01 (0.29) & remind.v.01 (0.17) & draw\_up.v.04 (0.14)\\
describe.v.02 & describe.v.01 (0.17) & kvetch.v.01 (0.29) & mislead.v.02 (0.91)\\
give.v.19 & give.v.03 (0.50) & give.v.03 (0.50) & give.v.01 (0.13)\\
balloon.v.02 & balloon.v.02 (1.00) &  (0.00) & bathe.v.01 (0.21)\\
dress.v.10 & dress.v.01 (0.29) & overcook.v.01 (0.26) & repair.v.01 (0.25)\\
ace.v.03 & ace.v.02 (0.29) & improve.v.01 (0.18) & dig.v.01 (0.20)\\
poach.v.02 & poach.v.01 (0.18) & catch.v.04 (0.18) & pour.v.01 (0.18)\\
hawk.v.02 & hawk.v.01 (0.44) & sign.v.05 (0.29) & meow.v.01 (0.19)\\
shuffle.v.02 & shuffle.v.01 (0.20) & braid.v.03 (0.17) & toss.v.03 (0.22)\\
bust.v.02 & bust.v.01 (0.33) & push.v.01 (0.46) & block.v.01 (0.44)\\
check.v.19 & check.v.01 (0.19) & recognize.v.04 (0.30) & draw\_up.v.04 (0.30)\\
plug.v.04 & plug.v.05 (0.25) & bewitch.v.01 (0.33) & search.v.01 (0.35)\\
ring.v.06 & ring.v.01 (0.13) & wave.v.01 (0.18) & write.v.07 (0.42)\\
bark.v.03 & bark.v.04 (0.17) &  (0.00) & decapitate.v.01 (0.59)\\
refresh.v.02 & refresh.v.01 (0.40) & leak.v.04 (0.26) & relax.v.01 (0.18)\\
take.v.27 & take.v.09 (0.45) & take.v.09 (0.45) & run.v.01 (0.78)\\
draw.v.07 & draw.v.06 (0.48) & draw.v.06 (0.48) & draw.v.13 (0.92)\\
order.v.05 & order.v.02 (0.71) & order.v.02 (0.71) & dial.v.02 (0.56)\\
cram.v.03 & cram.v.02 (0.15) & demolish.v.03 (0.62) & call.v.05 (0.12)\\
cram.v.02 & cram.v.01 (0.20) & tear.v.01 (0.83) & slice.v.03 (0.80)\\
challenge.v.02 & challenge.v.01 (0.15) & pick\_up.v.02 (0.36) & elect.v.01 (0.47)\\
moderate.v.03 & moderate.v.03 (1.00) & clear.v.24 (0.46) & grow.v.02 (0.17)\\
book.v.03 & book.v.02 (0.14) & marry.v.01 (0.55) & allow.v.04 (0.14)\\
solicit.v.03 & solicit.v.02 (0.25) & spy.v.02 (0.40) & sentence.v.01 (0.46)\\
hobble.v.03 & hobble.v.01 (0.25) & brush.v.01 (0.11) & trap.v.04 (0.16)\\
wax.v.03 & wax.v.01 (0.40) & shine.v.02 (0.19) & wake\_up.v.02 (0.76)\\
breach.v.02 & breach.v.01 (0.22) & leave.v.05 (0.18) & execute.v.03 (0.67)\\
swan.v.03 & swan.v.01 (0.25) & send.v.01 (0.70) & swim.v.01 (0.45)\\
\bottomrule
\end{tabular}
\end{table}

\begin{table}[hbtp]
\caption{Instances of the challenge set with modifiers with out-of-vocabulary concepts.}
\setlength{\tabcolsep}{9pt}
\label{tab:all_adjv}
\scriptsize
\begin{tabular}{rrrr}
\toprule
Gold  & LPS   & WID & TAX \\ 
\midrule
calm.a.02 & calm.a.01 (0.16) &  (0.00) &  (0.00)\\
rare.a.03 & rare.a.01 (0.42) & capable.a.05 (0.42) & large.a.01 (0.67)\\
firmly.r.02 & firmly.r.02 (1.00) & all\_of\_a\_sudden.r.01 (0.33) & comfortably.r.02 (0.47)\\
sturdy.a.03 & sturdy.a.02 (0.50) & dirty.a.01 (0.47) & dirty.a.01 (0.47)\\
short.a.02 & short.a.02 (1.00) & short.a.02 (1.00) & short.a.02 (1.00)\\
muscular.a.02 & muscular.a.01 (0.00) &  (0.00) & fat.a.01 (0.50)\\
hard.a.03 &  (0.00) &  (0.00) & jacket.n.01 (0.00)\\
grand.a.08 & grand.a.01 (0.50) & huge.a.01 (0.17) & huge.a.01 (0.17)\\
extended.a.03 & extended.a.01 (0.50) & long.a.02 (0.93) & long.a.01 (0.38)\\
gently.r.02 & gently.r.01 (0.32) & in\_public.r.01 (0.47) &  (0.50)\\
dry.a.02 & dry.a.01 (0.38) & surprised.a.01 (0.33) & weak.a.01 (0.40)\\
broken.a.07 & broken.a.03 (0.50) & international.a.01 (0.44) & right.a.02 (0.53)\\
special.a.04 & special.a.01 (0.00) &  (0.00) & bad.a.01 (0.18)\\
vicious.a.02 & vicious.a.02 (1.00) & fishy.a.02 (0.43) & suspicious.a.01 (0.47)\\
rugged.a.03 & rugged.a.01 (0.59) & dirty.a.01 (0.42) & upper.a.01 (0.16)\\
rather.r.04 & rather.r.02 (0.86) & rather.r.04 (1.00) & rather.r.04 (1.00)\\
sleazy.a.02 & sleazy.a.01 (0.50) & overweight.a.01 (0.53) & greasy.a.02 (0.42)\\
immature.a.05 & immature.a.01 (0.50) & tense.a.01 (0.20) & impenetrable.a.01 (0.57)\\
plumy.a.03 &  (0.00) &  (0.00) &  (0.00)\\
fairly.r.03 & fairly.r.02 (0.50) & quickly.r.02 (0.50) & seriously.r.01 (0.50)\\
unfledged.a.02 & unfledged.a.01 (0.50) & nuts.a.01 (0.17) & wounded.a.01 (0.22)\\
furious.a.02 & furious.a.02 (1.00) & angry.a.01 (0.96) & scared.a.01 (0.45)\\
uncontrollable.a.03 & uncontrollable.a.01 (0.50) & dirty.a.01 (0.27) & unemployed.a.01 (0.52)\\
smart.a.05 & smart.a.01 (0.35) & long.a.01 (0.62) & slow.a.03 (0.33)\\
horny.a.02 & horny.a.01 (0.50) & wild.a.02 (0.17) & ridiculous.a.02 (0.17)\\
kafkaesque.a.02 &  (0.00) &  (0.00) & armed.a.01 (0.20)\\
vivid.a.03 & vivid.a.02 (0.50) & tired\_of.a.01 (0.20) & excruciating.a.01 (0.42)\\
\bottomrule
\end{tabular}
\end{table}

\clearpage
\appendixsection{Hierarchy Reflection Score \label{app:HRS}}

Algorithm~\ref{code:CalculateScore} shows the calculation of Hierarchy Reflection Score (HRS-all). To be more detailed, in our four specificity levels, for the inequalities $d(0,1) < d(0,2)$, 
                                          $d(0,1) < d(0,3)$,
                                          $d(0,2) < d(0,3)$,
                                          $d(1,2) < d(0,2)$,
                                          $d(1,2) < d(1,3)$,
                                          $d(1,3) < d(2,3)$,
                                          $d(2,3) < d(0,3)$, HRS is $\frac{1}{7}$ if one of them is satisfied; HRS is $\frac{2}{7}$ if two of them is satisfied;\(\ldots\); HRS is $\frac{7}{7}$ if all of them are satisfied.

\begin{algorithm}
\small
\caption{Calculate Score for n Embeddings}
\label{code:CalculateScore}
\begin{algorithmic}[1]
\REQUIRE Embeddings $\text{emb}_0, \text{emb}_1, \ldots, \text{emb}_{n-1}$
\STATE $score \gets 0$
\STATE $totalComparisons \gets 0$
\STATE $d(p, q) \gets 1 - \cos(\text{emb}_p, \text{emb}_q)$

\FOR{$i \gets 0$ to $n-1$}
    \FOR{$j \gets 0$ to $n-1$}
        \IF{$i \neq j$}
            \FOR{$k \gets 0$ to $n-1$}
                \IF{$k \neq i \ \AND \  k \neq j$}
                    \STATE $totalComparisons \gets totalComparisons + 1$
                    \IF{$d(i, j) > d(k, j)$}
                        \STATE $score \gets score + 1$
                    \ENDIF
                \ENDIF
            \ENDFOR
        \ENDIF
    \ENDFOR
\ENDFOR

\RETURN $score / totalComparisons$
\end{algorithmic}
\end{algorithm}

\clearpage
\starttwocolumn
\bibliography{compling_style}
\end{document}